\documentclass{article}

\usepackage[hidelinks]{hyperref}
\usepackage{PRIMEarxiv}
\usepackage{orcidlink}
\usepackage{algpseudocode}
\usepackage{algorithm}
\usepackage{amsmath}

\usepackage{multirow}
\usepackage{float}
\usepackage[utf8]{inputenc} 
\usepackage[T1]{fontenc}    
\usepackage{hyperref}       
\usepackage{url}            
\usepackage{booktabs}       
\usepackage{amsfonts}       
\usepackage{fancyhdr}       
\usepackage{graphicx}       
\usepackage{authblk}
\usepackage{caption}
\title{Machine and Deep Learning Methods with Manual and Automatic Labelling for News Classification in Bangla Language
}

\author[1]{Istiak Ahmad\orcidlink{0000-0001-9914-4116}}
\author[1]{Fahad AlQurashi\orcidlink{0000-0002-7919-747X}}
\author[2,*]{Rashid Mehmood\orcidlink{0000-0002-4997-5322}}

\affil[1]{Department of Computer Science, Faculty of Computing and Information Technology, King Abdulaziz University, Jeddah 21589, Saudi Arabia}

\affil[2]{High Performance Computing Center, King Abdulaziz University, Jeddah 21589, Saudi Arabia}

\affil[*]{Corresponding author: RMehmood@kau.edu.sa}

\begin{document}
\maketitle

\begin{abstract}
Research in Natural Language Processing (NLP) has increasingly become important due to applications such as text classification, text mining, sentiment analysis, POS tagging, named entity recognition, textual entailment, and many others. This paper introduces several machine and deep learning methods with manual and automatic labelling for news classification in the Bangla language. We implemented several machine (ML) and deep learning (DL) algorithms. The ML algorithms are Logistic Regression (LR), Stochastic Gradient Descent (SGD), Support Vector Machine (SVM), Random Forest (RF), and K-Nearest Neighbour (KNN), used with Bag of Words (BoW), Term Frequency-Inverse Document Frequency (TF-IDF), and Doc2Vec embedding models. The DL algorithms are Long Short-Term Memory (LSTM), Bidirectional LSTM (BiLSTM), Gated Recurrent Unit (GRU), and Convolutional Neural Network (CNN), used with Word2vec, Glove, and FastText word embedding models. We develop automatic labelling methods using Latent Dirichlet Allocation (LDA) and investigate the performance of single-label and multi-label article classification methods. To investigate performance, we developed from scratch Potrika, the largest and the most extensive dataset for news classification in the Bangla language, comprising 185.51 million words and 12.57 million sentences contained in 664,880 news articles in eight distinct categories, curated from six popular online news portals in Bangladesh for the period 2014-2020. GRU and Fasttext with 91.83\% achieve the highest accuracy for manually-labelled data. For the automatic labelling case, KNN and Doc2Vec at 57.72\% and 75\% achieve the highest accuracy for single-label and multi-label data, respectively. The methods developed in this paper are expected to advance research in Bangla and other languages.
\end{abstract}


\section{Introduction}

The primary objective of text classification is to determine the class or sentiment of the unknown texts. We can define the problem as follows. Assume, we have n texts, $x = x_{1}, x_{2}, ..., x_{n}$, and each of them is assigned a category from a set of categorical values l, where $l = \{l_{1}, l_{2}, ... \}$. The training dataset is applied for generating a classification model, which relates the features to one of the class labels. The trained classification model can ascertain the unknown class from the text. Typically, texts are not tagged; we have to do so manually, which is the most time-consuming and challenging task. Additionally, without tagged texts, it’s complicated to develop a classification model. Text classification has made continuous success in many applications such as sentiment analysis~\cite{medhat2014sentiment}, information retrieval~\cite{dwivedi2016automatic}, information filtering, knowledge management, document summarization~\cite{cao2017improving}, spam mail detection~\cite{karim2019comprehensive}, recommended systems, and many others, which have become immense and boundless. 

About 238 million people speak Bangla natively or as a second language throughout the world (2021)~\cite{bangla_rank}. As a result, this language has carved out a niche for itself in different information-exchanging media. Bangla, with a large number of online newspapers, blogs, Wikipedia, eBooks, literature, and so on, may be considered to be following the NLP's action ground contest in the imminent future. Each day, a lot of events happening around the world, and some of those events become more trendy discussion topics for a certain time. Most of the news media are engaged in presenting the most popular events every time. Everyone desires to follow the most influential and frequently discussed events among a large number of events happening around us at a specific time. To get the most contemporary discussion topics and events, text analysis can automatically detect them more precisely and speedily. The research on text analysis. 

This paper introduces several machine and deep learning methods with manual and automatic labelling for news classification in the Bangla language. In the case of manual labelling, we implemented several machine (ML) and deep learning (DL) algorithms. The ML algorithms are Logistic Regression (LR), Stochastic Gradient Descent (SGD), Support Vector Machine (SVM), Random Forest (RF), and K-Nearest Neighbour (KNN), used with Bag of Words (BoW), Term Frequency-Inverse Document Frequency (TF-IDF), and Doc2Vec embedding models. The DL algorithms are Long Short-Term Memory (LSTM), Bidirectional LSTM (BiLSTM), Gated Recurrent Unit (GRU), and Convolutional Neural Network (CNN), used with Word2vec, Glove, and FastText word embedding models. 
To address the challenges related to the arduous task of manual labelling, we develop automatic labelling methods using Latent Dirichlet Allocation (LDA), an unsupervised topic modelling algorithm and investigate the performance of single-label and multi-label article classification methods.

We developed Potrika -- the largest and the most extensive dataset for news classification in the Bangla language -- comprising 185.51 million words and 12.57 million sentences contained in 664,880 news articles, and used it to investigate the proposed ML and DL methods~\cite{https://doi.org/10.48550/arxiv.2210.09389}. 
Potrika is a single-label news article textual dataset in the Bangla language curated for NLP research from six popular online news portals in Bangladesh (Jugantor, Jaijaidin, Ittefaq, Kaler Kontho, Inqilab, and Somoyer Alo) for the period 2014-2020. The articles are classified into eight distinct categories (National, Sports, International, Entertainment, Economy, Education, Politics, and Science \& Technology). 
GRU and Fasttext with 91.83\% achieve the highest accuracy for manually-labelled data. For the automatic labelling case, KNN and Doc2Vec at 57.72\% and 75\% achieve the highest accuracy for single-label and multi-label data, respectively. 
The lower performance for automatic-labelling-based classification is because it uses ML algorithms compared to the case of classification with manually-labelled data where the best performance was obtained using a DL algorithm. We will extend our work in the future to include DL methods for automatic labelling.

The NLP methods developed in this paper and the techniques for their extensive analyses are expected to advance research in Bangla and other languages. 

\paragraph{\textbf{Hardware and Software:}} We use the Quadro RTX-6000 GPU, which has 4608 CUDA Parallel-Processing Cores, 576 tensor cores, and 72 RT Cores. The GPU memory is 24 GB of GDDR6. We use Python as the programming language along with machine and deep learning libraries like Tensorflow, Keras, Scikit-Learn, Gensim, etc. We use data visualization libraries like Seaborn and Matplotlib to visualize the evaluation results.

The following is how the paper is structured: Section~\ref{sec.literatureReview} describes the literature review of text classification, followed by the Bangla text classification (Sections~\ref{sec.literatureReview.sentiment} to~\ref{sec.literatureReview.topic}) and other language text classification (Section~\ref{sec.literatureReview.others}). Section~\ref{sec.methodology} discusses the proposed methodologies of our research including, the overview of methodology and framework architecture~\ref{sec.methodology.overview}, dataset~\ref{sec.dataset}, preprocessing~\ref{sec.DataPreprocessing}, feature extraction~\ref{sec.featureextraction}, and word embedding techniques~\ref{sec.WordEmbeddingTechniques}. Machine and deep learning methods for manual labelling are described in the Sections~\ref{sec.ML} and~\ref{sec.DL}. Section~\ref{sec.automatic.single} explains the methodology for creating the automatically labelled dataset using the unsupervised topic modeling method, and Section~\ref{sec.automatic.multi} discusses the methodology of multi-label news article classification. Subsequently, the results of all proposed methods are discussed in Sections~\ref{sec.result.manual.ml} and~\ref{sec.automatic.result}, which depict the manual labeling, and automatic single labeling with multi-labeling news article classification results, respectively. Section~\ref{sec.discussion} describe the discussion of the paper. Finally, in Section~\ref{sec.conclusion}, we conclude with recommendations for further research. 

\section{Literature Review}\label{sec.literatureReview}
To address text classification~\cite{minaee2020deep}, several machine and deep learning-based approaches have been introduced. In this part, we will go through how to classify Bangla text using sentiment analysis, multi-domain, and topic modeling methods. We also go through several methods for classifying other language-related texts.

\subsection{Sentiment Classification}\label{sec.literatureReview.sentiment}
The core idea of sentiment analysis, or opinion mining, is to analyses the addressed text, if the text expression holds positive, negative, or neutral meaning.
For sentiment analysis in Bangla, TF-IDF was applied to a small dataset using machine learning algorithms (see~\cite{nabi2016detecting,mahtab2018sentiment,tabassum2019design}). The word embedding method named word2vec was proposed by~\cite{al2017sentiment} for Bangla sentiment analysis based on the Bangla comments. Bangla tweet data is also used for sentiment analysis. For example, Asimuzzaman et al.~\cite{asimuzzaman2017sentiment} used an adaptive neuro-fuzzy system for Bangla tweet classification. For sentiment detection, Hasan et al.~\cite{hasan2014sentiment} proposed WordNet and SentiWordNet as tools but the major limitation of this research was proposed tools were developed specifically for English. Tuhin et al.~\cite{tuhin2019automated} predicted six individual emotions using ML algorithms such as SVM and NB. Further, NB, DT, KNN, SVM, and K-means clustering were also used by Rahman et al.~\cite{rahman2019comparison} to predict some basic emotions from the text. In addition, mutual information-based feature selection methods and the multi NB algorithm proposed by Paul et al.~\cite{paul2016sentiment} for predicting sentiment polarity. N-gram and SVM based Bangla sentiment analysis proposed by Taher et al.~\cite{taher2018n}. A popular English tool called VADER was proposed by Amin et al.~\cite{amin2019bengali} to predict Bangla sentiment.

A deep learning-based algorithm named LSTM was proposed by Hassan et al.~\cite{hassan2016sentiment} for sentiment analysis, where they used 10k Bangla and romanized Bangla text (BRBT) dataset with binary and a categorical cross-entropy loss function. Furthermore, the CNN-based method was proposed by Alam et al.~\cite{alam2017sentiment}.

\subsection{Multi-domain Text Classification}\label{sec.literatureReview.multi}
Alam et al.~\cite{alam2018bard} presented a new dataset for Bengali news articles which contains about 350K articles in five categories (State, Economy, International, Entertainment, and Sports). In their dataset, 65\% of the data is labelled as State and 13.5\%, 8.5\%, 8\% and 5\% are labelled as Sports, International, Entertainment, and Economy respectively. They have applied machine learning algorithms with two word embedding techniques such as Word2Vec and TFIDF. In another study, the Word2vec embedding model was implemented with KNN and SVM classification algorithms by Ahmed et al.~\cite{ahmad2016bengali} for news document classification. A classification technique based on cosine similarity and Euclidean distance based on a set of 1000 documents was recommended by Dhar et al.~\cite{dhar2017classification}. They measure the {\ensuremath{\beta}}0 threshold using the 90th percentile formula for both the distance measures and calculate the score based on the distance. In another research, the dimensional reduction technique with TFIDF (40\% of TF) was developed by Dhar et al.~\cite{dhar2018application} where they used a total of 1960 Bangla text documents from five categories (Sports, Business, Science, Medical, and State) with  632,924 tokens and applied machine learning algorithms. The classification algorithm LIBLINEAR achieved the highest accuracy. For 40 thousand news samples divided into 12 categories, Mojumder et al.~\cite{mojumder2020study} suggested DL algorithms such as BiLSTM, CNN, and convolutional BiLSTM, and fastText as word embedding techniques. The Bangla article classification based on transformers was proposed by Alam et al.~\cite{alam2020banglaTransformers}. In this study, they used multilingual transformer models to classify Bangla text in several areas.

\subsection{Topic Modeling-based Text Classification} \label{sec.literatureReview.topic}
Scarce research has been performed to classify Bangla text using topic modeling. Helal and Mouhoub~\cite{helal2018topic} find the key topics in the Bangla news corpus using LDA with a bigram model and classify them by applying similarity measures. They evaluated the proposed model using the LDA and Doc2Vec models and compared the similarity scores. They point out that in some specific news articles, the LDA performance is better than the Doc2Vec model. Alam et al.~\cite{alam2020bangla} also proposed an LDA-based topic modeling algorithm using 70k Bangla news articles. They detect 5 distinct news article topics (National, Sports, International, Technology, and Economy) and another topic called ‘others’ which exclude the following distinct topics.  

Most of the above research work has been done on machine learning algorithms with small datasets, but there has been remarkably little works on deep learning algorithms for Bangla article classification because there is no comprehensive dataset for Bangla article classification. 

\subsection{Text Classification for Other Languages}\label{sec.literatureReview.others}

This section provides an overview of different text classification methods for other languages. Shaw et al.~\cite{shah2020comparative} implemented ML techniques including random forest, KNN, and logistic regression to classify the news into five categories (Entertainment, Business, Politics, Sports and Technology) based on the BBC news dataset. In terms of efficiency among these algorithms, it turned out that logistic regression has better performance for all the categories. Another research on three machine learning algorithms, namely SVM, neural network, and decision tree, has been done by Raychaudhuri et al.~\cite{raychaudhuri2016comparative} for text classification. The authors used the UCI dataset on US congressional voting that consists of 16 features, 435 instance examples, 335 examples of the training dataset, 50 examples of the testing dataset, and 50 examples of the validation dataset. They used variable C, which controls the training error. When C=1, SVM performed better than the neural network and when C=1000, the neural network performed better. The outcome also revealed that a fully grown decision tree produced better results than a smaller decision tree. 

The data augmentation technique is most popular in computer vision research when the amount of data is small or imbalanced. Recently, the text data augmentation technique is noted for small text datasets. Wei and Zou~\cite{wei2019eda} proposed this technique to increase the text classification performance. The following operations are proposed for data augmentation: synonym replacements, random insertion of synonyms of a word, randomly swapping two words positions in a sentence, and randomly removing words in a sentence.

Recently, the attention mechanism has become an efficient approach to determine the important erudition to achieve excellent outcomes. Numerous studies have been carried out on attention mechanisms and architecture. For text classification, several novel methods are also proposed~\cite{zhou2016attention, du2017convolutional, liu2019bidirectional, li2020bidirectional}. An attention-based LSTM network was proposed by Zhou et al.~\cite{zhou2016attention} to classify cross-lingual sentiments, where they used English and Chinese as the source and target languages, respectively. A Convolutional-Recurrent Attention Network (CRAN) was proposed by Du et al.~\cite{du2017convolutional}. Their proposed architecture includes a text encoder using RNN, and an attention extractor using CNN. The experimental result shows that the model effectively extracts the salient parts from sentences along with improving the sentence classification performance. Liu et al.~\cite{liu2019bidirectional} proffered attention-based convolution layer and BiLSTM architecture, where the attention mechanism provides a focus for the hidden layers output. The BiLSTM is used to extract both previous and following context, while the convolutional layer retrieves the higher-level phrase from the word embedding vectors. Their experimental results get comparable results for all the benchmark datasets.

The state-of-the-art graph-based neural network methods for text classification have been gaining increasing attention recently. A text graph convolutional network (TextGCN) was proposed by Yao et al.~\cite{yao2019graph}, which is more notable for its small training corpus for text classification. To learn the TextGCN for the corpus, word co-occurrence and the relation between the word document based single text graph was developed.
Another tensor graph convolutional network (TensorGCN) has been proposed by Liu et al.~\cite{liu2020tensor}. They develop the text graph tensor based on semantic, syntactic, and sequential contextual information. After that, two types of propagation learning are performed on the text graph tensor called intra-graph propagation to aggregate information from neighboring nodes and inter-graph propagation to tune heterogeneous information between graphs. 

Capsule network is another state-of-the-art method for text classification that is inherent to CNNs. Several studies based on the capsule network have been conducted~\cite{yang2019investigating, kim2020text, jain2020deep}. In capsule networks, capsules are locally invariant groups that learn to recognize the presence of visual entities and encode their characteristics into vectors. It also requires a nonlinear function called squashing, whereas neurons in a CNN act independently. However, equivariance and dynamic routing are the two most essential characteristics of Capsule Networks that distinguish them from standard Neural Networks. A Capsule network with dynamic and static routing based text classification methods was proposed by Kim et al.~\cite{kim2020text}.  Static routing achieved higher accuracy than dynamic routing.
Yang et al.~\cite{yang2019investigating} introduced a cross-domain capsule network and illustrated the transfer learning applications for single-label to multi-label text classification and cross-domain sentiment classification. An attention mechanism-based capsule network system called Deep Refinement was suggested by Jain et al.~\cite{jain2020deep}. Their proposed method achieved 96\% accuracy for text classification compared to BiLSTM, SVM, and C-BiLSTM for the Quora insincere question dataset.

Traditional text classification techniques use manually labelled datasets that are monotonous and time-consuming. Recently, a few dataless text classification techniques, for example, the Laplacian seed word topic model (LapSWTM)~\cite{li2018dataless}, and seed-guided multi-label topic model (SMTM)~\cite{zha2019multi} have recently been proposed to solve this challenge. Anantharaman et al.~\cite{anantharaman2019performance} proposed large and short text classification non-negative matrix factorization, LDA, and LSA (latent semantic analysis). LSA with TFIDF was proposed by Neogi et al.~\cite{neogi2020topic} for text classification. To increase the accuracy, they used entropy. A self-training LDA based semi-supervised text classification method was proposed by pavlinek et al.~\cite{pavlinek2017text} for text classification.

\subsection{Research Gap, Novelty, and Contributions} \label{sec.literatureReview.Gap}
Text datasets, often known as corpora, are used to study linguistic phenomena including text classification, morphological structure, word sense disambiguation, language evolution over time, and spelling checking. The quality and amount of the corpus have a big impact on the research output. A well-structured, comprehensive corpus can yield far superior study results. In comparison to the English language, there has been inadequate study done due to the paucity of the Bangla corpus and the complicated grammatical structure. In this paper, our contributions are as follows:

\begin{itemize}
	\item We are the first to use a comprehensive Bangla newspaper article dataset called Potrika~\cite{https://doi.org/10.48550/arxiv.2210.09389,istiakdata2021} to classify eight distinct news article classes, including Education, Entertainment, Sports, Politics, National, International, Economy, and Science \& Technology. 
	\item We implement both machine learning (ML) including logistic regression, SGD, SVM, RF and KNN algorithms, and deep learning (DL) including CNN, LSTM, BiLSTM, and GRU algorithms for single label news article classification. We perform BOW, TFIDF, and Doc2Vec word embedding models for ML algorithms. For DL algorithms, we apply word embedding models such as word2vec, glove, and fasttext that were developed based on the Potrika dataset. These word embedding models are not only valuable for news article classification but also for other NLP tasks like text summarization, named entity recognition, Bangla automatic word prediction, question-answering systems, etc. Further, we evaluate and scrutinise the results for both cases.
	\item Manual labeling is the most difficult and time-consuming task for classification datasets. In the following paper, we investigate the possibility of using the topic modeling algorithm to automatically label the news article dataset and compare the performance of the automatically labelled dataset with that of the manually labelled dataset. Additionally, we also develop another multi-label dataset based on the automatic label dataset and evaluate the multi-label news article classification's performance. 
\end{itemize}

The NLP work proposed in this paper builds on our earlier NLP works applied to several sectors and multiple languages including transportation~\cite{alomari2020iktishaf,alomari2021iktishaf+,ahmad2022deep}, healthcare~\cite{Alotaibi2019,su14063313}, education~\cite{10.3389/frsc.2022.751681, 10.3389/frsc.2022.871171}, and smart cities~\cite{alomari2021covid, suma2017enabling, suma2020automatic,su142013534}. We expect that this paper will significantly increase the impact of our work particularly in the Bangla language.

\section{Methodology and Design} \label{sec.methodology}

In this section, we describe our methodology for the research presented in this paper. We begin with an overview of our methodology in Section~\ref{sec.methodology.overview} followed by a description of the dataset in Section~\ref{sec.dataset}. Data preprocessing and feature extraction are explained in Sections~\ref{sec.DataPreprocessing} and~\ref{sec.featureextraction}. Word Embedding Models are discussed in Section~\ref{sec.WordEmbeddingTechniques}. Machine learning and deep learning techniques are discussed in Section~\ref{sec.ML} and~\ref{sec.DL}, respectively. 
Subsequently,  we explain our methodology for automatically creating labels for the news items. The details of automatic labeling with single labels are provided in Section~\ref{sec.automatic.single} and the details of marking news items with multiple labels are given in Section~\ref{sec.automatic.multi}.

\begin{figure}[!htb]
	\centering
	\includegraphics[scale=0.10]{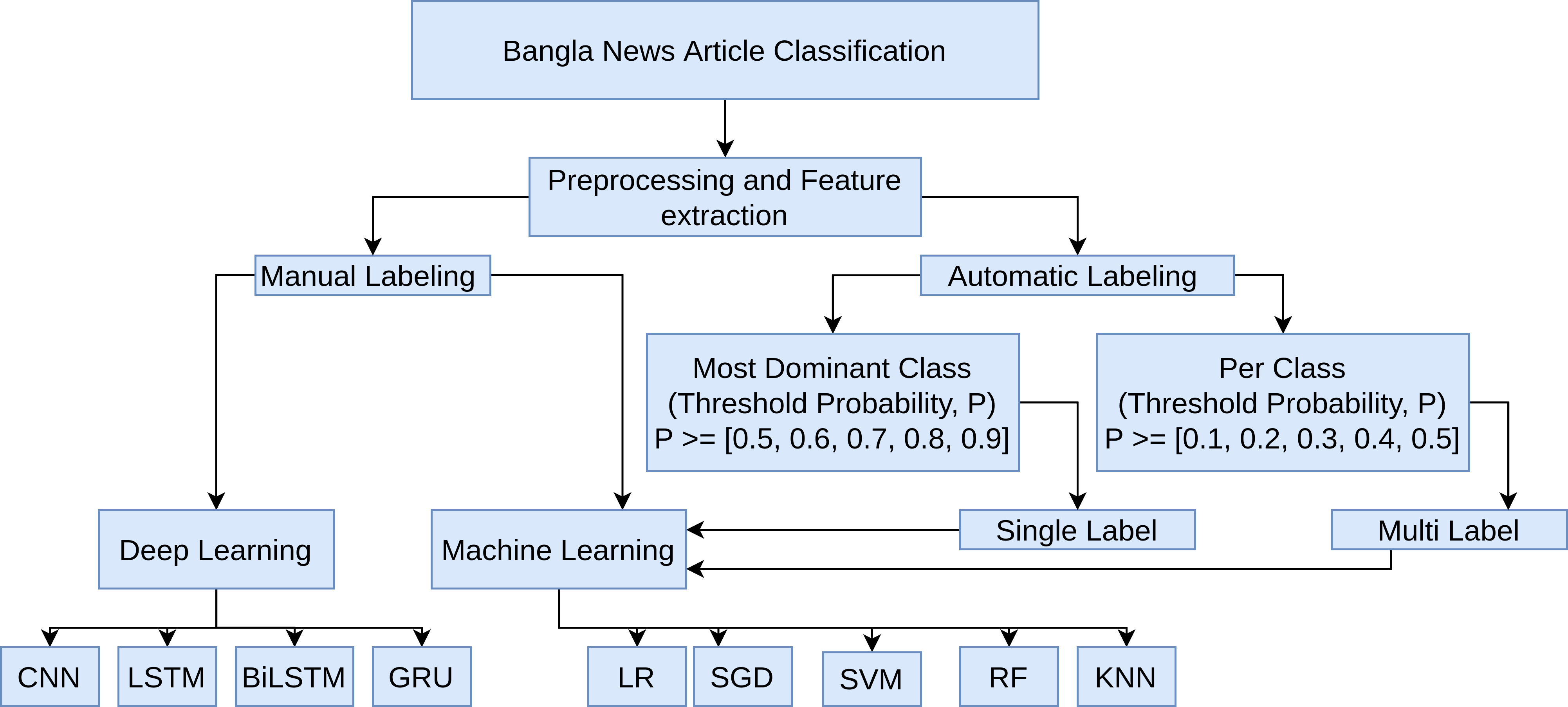}
	\captionsetup{justification=centering}
	\caption{System Process for Article Classification Overview}
	\label{fig.methodology.overview}
\end{figure}

\subsection{Methodology Overview}\label{sec.methodology.overview}
\begin{figure}[!htb]
		\centering
		\includegraphics[scale=0.10]{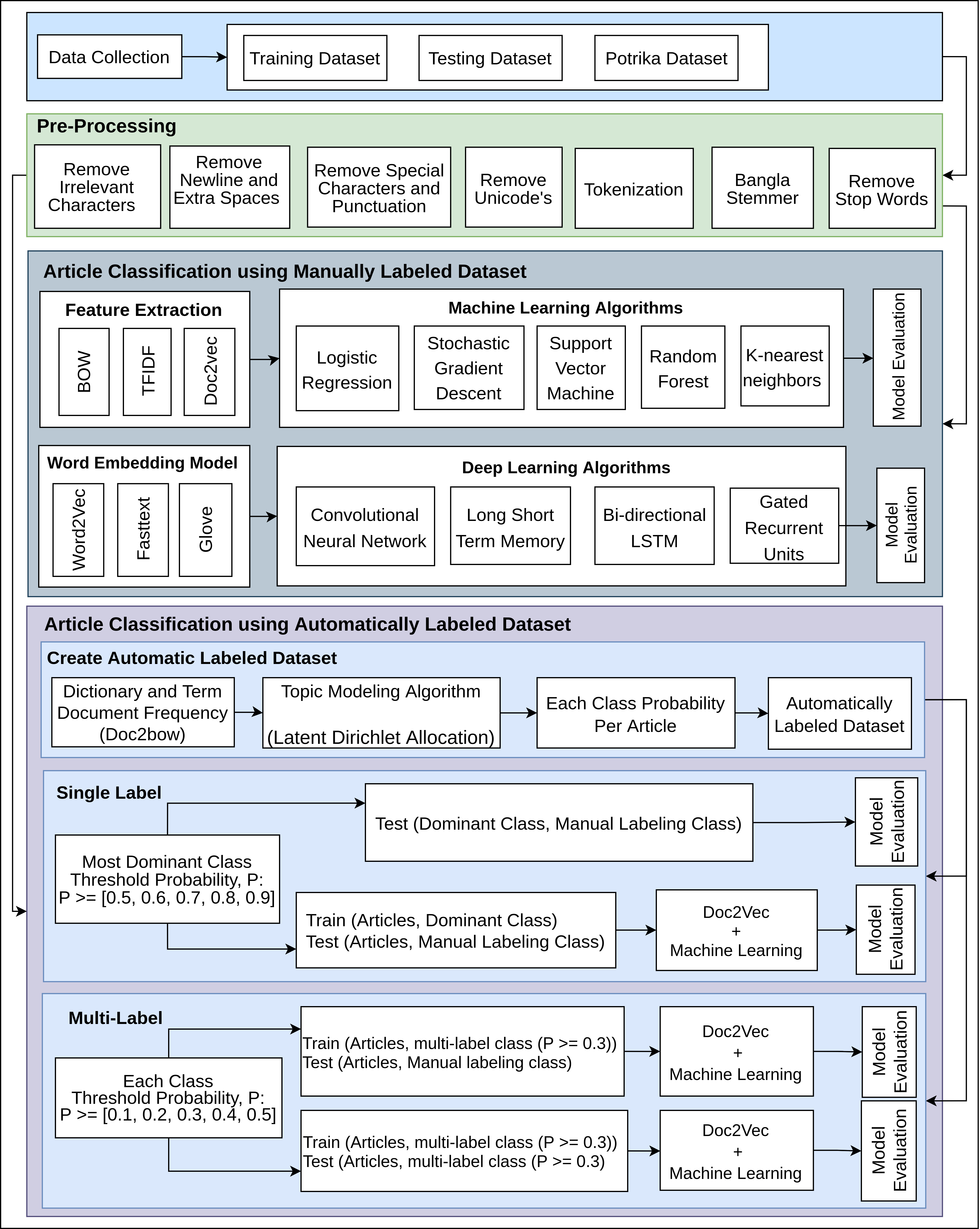}
		\captionsetup{justification=centering}
		\caption{A Detailed System Process for Article Classification}
		\label{fig.methodology.details}
\end{figure}

As mentioned earlier, the aim of the paper is to investigate the performance of machine and deep learning-based news classification in the Bangla language using manual and automatic labeling of news documents. Towards this end, we explore, firstly, the classification of manually labelled news data using machine and deep learning algorithms and, secondly, the classification of automatically labelled news items using single label and multi-label approaches. The automatic labeling is done using topic modeling. The overview and detailed architecture of our methodology are depicted in Figure~\ref{fig.methodology.overview} and Figure~\ref{fig.methodology.details} and its algorithmic flow is provided in Algorithm~\ref{algo.mainAlgo}.

\begin{algorithm}[h]
	\caption{Master Algorithm}
	\textbf{Input:} $Read trainDF, testDF, potrikaDF$ \\
	\textbf{Output:} $Evaluation of news article classification$
	\begin{algorithmic}[1]
		\State $\textit{cleanTrText, cleanTsText, cleanText} \gets \textit{preprocessing (trainDF, testDF, potrikaDF)}$
		\State $\textit{evaluation} \gets \textit{man\_ML(cleanTrText, trainDF.class, cleanTsText, testDF.class)}$
		\State $\textit{w2vmodel, glovemodel, fasttextmodel} \gets \textit{getWordEmbeddingModels(cleanText)}$
		\State $\textit{evaluation} \gets \textit{man\_DL(w2vmodel, glovemodel, fasttextmodel)}$
		\State $\textit{evaluation, autoLabelingDF} \gets \textit{auto\_singleLabel(trainDF, testDF)}$
		\State $\textit{evaluation} \gets \textit{auto\_multiLabel(autoLabelingDF)}$
	\end{algorithmic}
	\label{algo.mainAlgo}
\end{algorithm}

\subsection{Dataset} \label{sec.dataset}

\begin{figure}[!htb]
		\centering
		\includegraphics[scale=0.45]{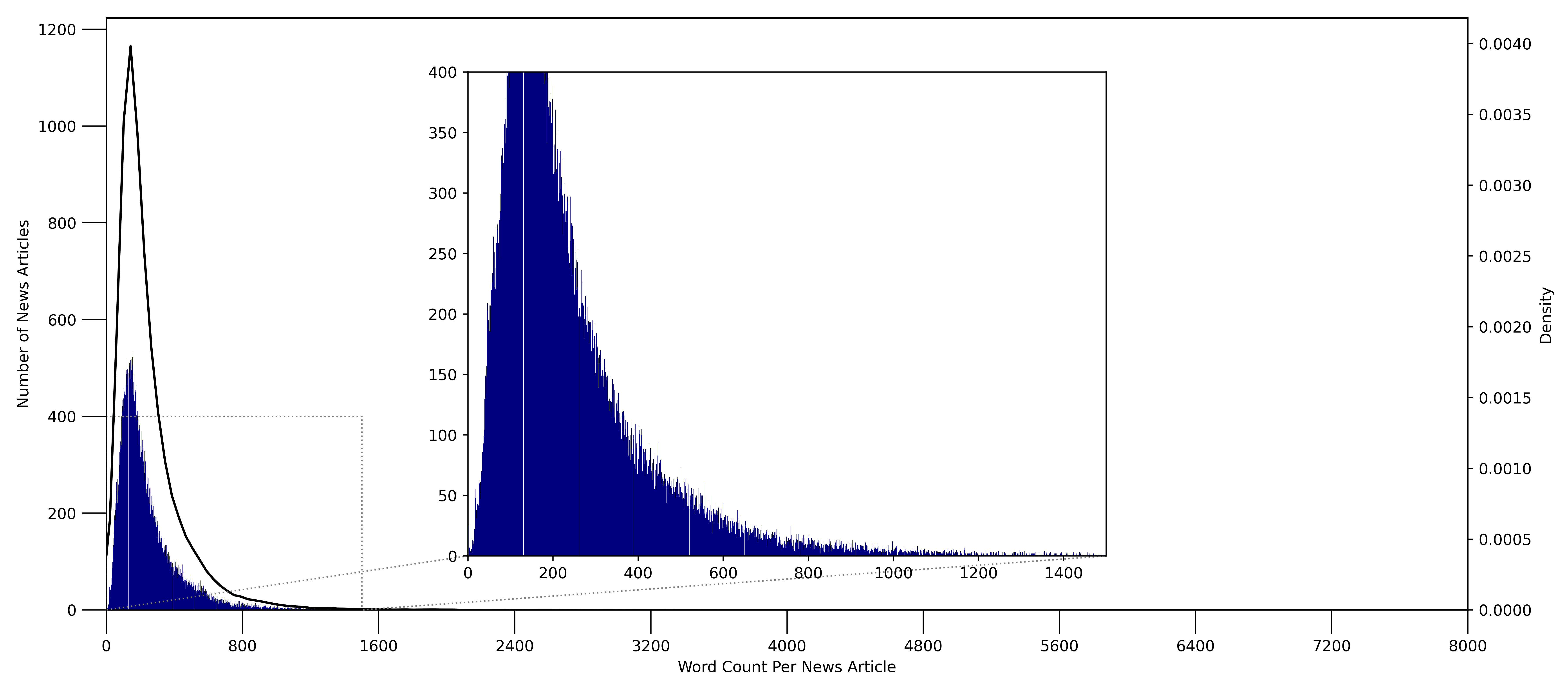}
		\captionsetup{justification=centering}
		\caption{Distribution of Document Word Counts}
		\label{fig.word_count}
\end{figure}

We used Potrika (for details, see~\cite{https://doi.org/10.48550/arxiv.2210.09389,istiakdata2021}), a large single-labelled Bangla News article dataset derived from six popular online news portals, including Jugantor, Jaijaidin, Ittefaq, Kaler Kontho, Inqilab, and Somoyer Alo, and divided into eight classes: National, Sports, International, Economy, Education, Politics, and Science \& Technology. The dataset has five columns for each class: news article, class, headline, publish date, and news source. Over 665K news articles, 12.5 million sentences, and 185.5 million words are included in the Potrika dataset. We used the Potrika dataset for word embedding models, and a total of 120K articles for text classification, including 100K for training and 20K for testing. Each class has 12.5K and 2.5K articles in the training and testing sets, respectively. The total amount of words each document/article vs the total number of documents/articles is depicted in Figure~\ref{fig.word_count}. 

Table \ref{tab.categoryShortform} depicts the short form of each category used in the result section.
\begin{table}[!htb]
	\centering
	\caption{Short Form of Each Category}
	\label{tab.categoryShortform}
	\begin{tabular}{|c|c|}
		\hline 
		Sports & SP \\
		\hline 
		National & NA \\ 
		\hline 
		Economy & EC \\
		\hline 
		Entertainment & EN \\ 
		\hline 
		Politics & PO \\
		\hline 
		International & IN \\
		\hline 
		Education & ED \\
		\hline 
		Science\&Technology & ST \\ 
		\hline 
	\end{tabular}
\end{table}

\subsection{Pre-Processing} \label{sec.DataPreprocessing}
Most of the texts include several irrelevant words, such as stop words, misspellings, unrecognized characters, Unicode, and so on, which might have inimical impacts on research. This section briefly explains the pre-processing technique to clean the texts. We have used similar pre-processing techniques for all the proposed methods. Moreover, we also provide an algorithm~\ref{algo.preprocess} to illustrate the pre-processing techniques.

\begin{algorithm}[H]
	\caption{Pre-processing}
	\textbf{Input:} $\textit{trainDF, testDF, and potrikaDF}$ \\
	\textbf{Output:} $\textit{cleanTrText, cleanTsText, cleanText}$			
	\begin{algorithmic}[1]
		\State $cleandataset \gets $initialize with empty list
		\For{$i = $0 $ to $ datasetLength}
		\State $\textit{cleanArticle} \gets $\textit{article[i]}
		\State $cleanArticle \gets $\textit{re.sub('[a-zA-Z0-9]', '', cleanArticle)}
		\State $\textit{cleanArticle} \gets $\textit{cleanArticle.replace('\textbackslash n', '')}
		\For {$x = $0 $ to $ length of bangladigitlist}
		\State $\textit{cleanArticle} \gets $\textit{cleanArticle.replace(x, ' ')}
		\EndFor
		\State $\textit{cleanArticle} \gets $\textit{translate(str.maketrans('', '', string.punctuation))}
		\State $\textit{cleanArticle} \gets $\textit{cleanArticle.replace(u'\textbackslash uf06c', '')}
		\State $\textit{cleanArticle} \gets $\textit{re.sub(r'\textbackslash x9d', r'', cleanArticle)}
		\State $\textit{cleanArticle} \gets $\textit{re.sub('  *', ' ', cleanArticle)}
		\State $\textit{cleanArticle} \gets $\textit{cleanArticle.split()}
		\State $bs \gets $\textit{stemmer.BanglaStemmer()}
		\State $\textit{cleanArticle} \gets $\textit{[bs.stem(word) for word in cleanArticle]}
		\State $\textit{cleanArticle} \gets $\textit{[word for word in cleanArticle if not word in bangla\_stopwords]}
		\State $\textit{cleanArticle} \gets $\textit{' '.join(cleanArticle)}
		\State $\textit{cleandataset.append(cleanArticle)}$
		\EndFor		
	\end{algorithmic}
	\label{algo.preprocess}
\end{algorithm}

\begin{figure}[!htb]
	\centering
	\includegraphics[scale=0.9]{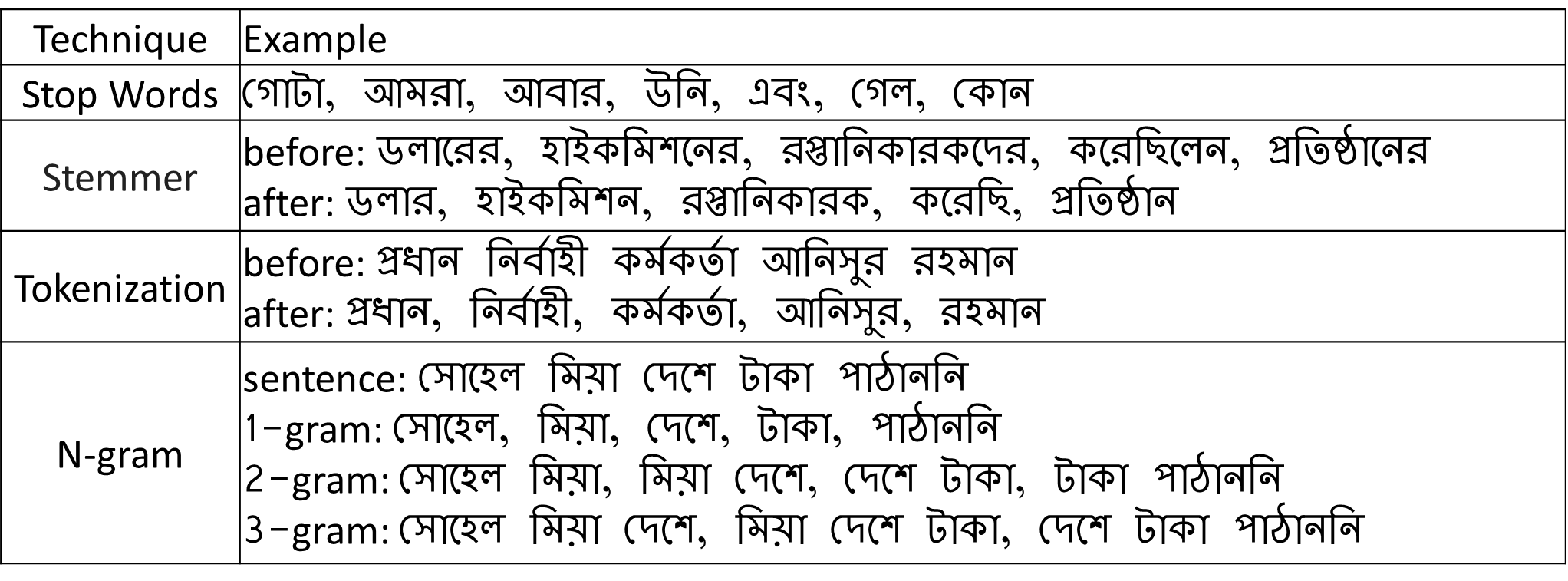}
	\captionsetup{justification=centering}
	\caption{Text Pre-Processing Techniques Example}
	\label{fig.preprocess}
\end{figure}

There are numerous optional Bangla words that do not have major consequences in classification algorithms. We manually generate a list of 542 stop words based on the significance of these words and remove from the texts (see Figure~\ref{fig.preprocess}). 

An enormous number of special and punctuation characters (i.e., \verb|“!#$%&'()*+?@|) are included in the texts. This punctuation and special characters are removed.

In the texts, we found some Unicode characters including \verb|‘\uf06c’|, \verb|‘\u200c’|, \verb|'\u09e5'|, \verb|'\x9d'|, etc., which we removed from the text to improve the accuracy of classification models.

A single word might appear in diverse forms in a document, all of which have the same semantic meaning. Stemming refers to consolidating different versions of words into the same feature location. In our research, we used python Bangla stemmer (version 1.0) to margin the same words. Figure~\ref{fig.preprocess} shows some examples before and after applying the Bangla stemmer.

Tokenization is a preprocessing technique that distributes texts into tokens, which can be words, phrases, symbols, or other significant components. To process the tokenization of the texts, text classification requires a parser. Figure~\ref{fig.preprocess} shows some examples before and after applying tokenization.

The feature extraction approach has been used by a number of researchers to deal with the problem of losing syntactic and semantic correlations inside words. We used the n-gram technique (1-gram, 2-gram, 3-gram) to address the syntactic problem. We used the 1-gram technique for manual labeling and the 3-gram technique for automatic labeling. Figure~\ref{fig.preprocess} shows some examples of n-gram.


\subsection{Feature Extraction}\label{sec.featureextraction}
Feature extraction or document representation is the most prominent method for enhancing the performance of text classification, since it enables us to determine which features are most relevant for the intended classification job. It should be done more accurately because of the vast dimensional of text features and the presence of irrelevant or noisy data. Usually, texts are unstructured and unorganized data that needs to be transformed into structured data using mathematical modeling as part of a classifier. Word embedding techniques are well-known feature extraction approaches that are covered in Section~\ref{sec.WordEmbeddingTechniques}.

\subsection{Word Embedding Models}  \label{sec.WordEmbeddingTechniques}
Several word embedding methods are discussed in this section, including BOW, TFIDF, Doc2Vec, word2vec, fasttext, and glove. BOW counts the total occurrence of a word in a document, and TFIDF assesses the significance of a word in a document, not its frequency. Word2Vec trains neural networks on documents and outputs a vector for each word. The embedding result catches whether words appear in similar contexts. The word2vec skip-gram model seeks to obtain co-occurrence one window at a time, whereas the glove seeks to obtain the counts of overall statistics on how often it appears. Both word2vec and glove train on the smallest units of words. Fasttext is a word2vec extension that treats each word as a set of n-grams. The following approach enables embedding to be trained on a smaller corpus and then generalized to unknown or rare words. A brief discussion of these word embedding methods is given below. 
We developed word embedding models for deep learning-based news article classification using the Potrika dataset. The procedure is depicted by the algorithm \ref{algo.word_embedding}.
\begin{algorithm}[H]
	\caption{Word Embedding Model}
	\textbf{Input:} $\textit{cleanText}$ \\
	\textbf{Output:} $\textit{Word Embedding Models}$
	\begin{algorithmic}[1]
		\State $\textit{w2vmodel} \gets \textit{word2vecmodel}(\textit{cleanText})$
		\State $\textit{glovemodel} \gets \textit{glovemodel}(\textit{cleanText})$
		\State $\textit{fasttextmodel} \gets \textit{fasttextmodel}(\textit{cleanText})$
	\end{algorithmic}
	\label{algo.word_embedding}
\end{algorithm}

\subsubsection{Bag of Words (BOW)}
The most basic type of numerical text description is the BOW model. To produce an unordered list of words without syntactic, semantic, or POS labeling, extract just the uni-gram words from the BOW. The document is represented by these groupings of words. There are various disadvantages to this strategy. If the new sentences contain new words, then the vocabulary dimension and vector length will grow. Moreover, multiple 0s may occur in the vectors, resulting in a sparse matrix that must be avoided.

\subsubsection{Term Frequency Inverse Document Frequency (TFIDF)}
A statistical metric for determining the importance of a word in a corpus or collection of documents is the TFIDF model. The significance of a word increases proportionally to its occurrence in the document, which remains offset by its recurrence in the corpus. We can divide TFIDF into two segments: term frequency (TF) and inverse document frequency (IDF). TF denotes a metric that measures the occurrence of a term in the current document. Since each document varies in length, a term may occur more frequently in large documents than in diminutive documents. For normalization, the TF is often divided by the length of the document. IDF denotes a metric that measures how rarely the word occurs in all documents. Equation \ref{eq.tfidf.1}, \ref{eq.tfidf.2}, and \ref{eq.tfidf.3} describe the formula of TFIDF method, where {\ensuremath{\zeta}} = number of documents in total, {\ensuremath{\beta}} = total amount of words in the document, {\ensuremath{\alpha}} = number of times a word w occurs in a document, and {\ensuremath{\delta}} = number of documents in which the word w occurs. \\

\begin{equation}\label{eq.tfidf.1}
TF_{w} = \frac{{\ensuremath{\alpha}}}{{\ensuremath{\beta}}}
\end{equation}
\begin{equation}\label{eq.tfidf.2}
IDF_{w} = log\frac{{\ensuremath{\zeta}}}{{\ensuremath{\delta}}}
\end{equation}
\begin{equation}\label{eq.tfidf.3}
TF\_IDF_{w} = {TF_{w} \times IDF_{w}}
\end{equation}

\subsubsection{Doc2Vec}
Doc2Vec is an extension of the word2vec (CBOW) model introduced by Mikilov and Le \cite{le2014distributed}. The word2vec model runs similarity queries to predict the subsequent words, whereas in the Doc2Vec model, we tag the document with extra tag vectors called document unique. When the word vectors W are trained, the document vector D is also trained, and it retains a numeric representation of the document at the conclusion of the training. The model is performing as memory that recalls what is lacking from the present context or the paragraph's topic. This model is also known as paragraph vector with distributed memory. The execution time for Doc2Vec is about 2.42 hours. There is a total of 664,883 tokens and 368,894 vocabularies in the Doc2Vec model. 

\subsubsection{Word2vec}
The word2vec was mentioned as an enhanced word embedding design by T. Mikolov et al.~\cite{mikolov2013distributed, mikolov2013efficient}. This approach used deep neural networks with two hidden layers, continuous bag-of-words (CBOW), and the continuous skip-gram model to create a high-dimensional vector per word and retain syntactic and semantic information of sentences. For the target phrase, the CBOW model is represented by numerous words. For instance, the context terms "aeroplane" and "military" for the target phrase "air-force." The continuous Skip-gram model, on the other hand, aims to maximize a word's classification depending on another word in the same phrase. The application of word2vec is enormous in deep learning, such as text classification, language modeling, question and answer systems, machine translations, image captioning, speech recognition, and so on. The execution time for word2vec is about 4 hours. There is a total of 664,883 tokens and 368,894 vocabularies in the word2vec model. 

\subsubsection{Glove}
Global Vectors (GloVe)~\cite{pennington2014glove} is a word embedding technique that is very comparable to the word2vec method. Here, each word is presented by a high-dimensional vector and trained on the surrounding words in a large corpus. Glove predicts surrounding words by using dynamic logistic regression to maximize the likelihood of occurrence of a context word given a core word. Other pre-trained word vectorizations with 100, 200, and 300 dimensions are available from the glove model, which has been trained on larger corpora. However, this pre-trained model is not suitable for the Bangla language. We trained the Glove model for the Potrika dataset for 300 dimensions. In the vocabulary building, 123,133,606 tokens are processed and total unique words 1,165,377 with 3 minimum occurrences accepted, and the total vocabulary size is 368,894. There are a total of 30 epochs and the execution time is 2.3 hours.

\subsubsection{Fasttext}
FastText~\cite{joulin2016fasttext} is an extension of the word2vec model that encodes each word as an n-gram of characters rather than learning vectors for words directly. It supports the training of CBOW or Skip-gram models using negative sampling, softmax or hierarchical softmax loss functions, and in terms of word representations and sentence classification, it performs admirably, especially for rare words, by utilizing character-level information. We trained a fasttext model with 300 dimensions, and the execution time was about 8.43 hours. There is a total of 664,883 tokens and 368,894 vocabularies in the fasttext model. 

\subsection{Machine Learning Algorithms}\label{sec.ML}
Before applying ML algorithms, we took the training and testing datasets separately and preprocessed both datasets to create a clean dataset. We discussed the preprocessing techniques in Section~\ref{sec.DataPreprocessing}. After preprocessing the dataset, we applied three-word embedding techniques~\ref{sec.WordEmbeddingTechniques} including BOW, TFIDF, and Doc2Vec for feature extraction. We used ML algorithms named RF, LR, KNN, SGD, and SVM for the news article classification. Algorithm~\ref{algo.manual_ML} shows the process for news article classification using machine learning algorithms.

\begin{algorithm}[H]
	\caption{News Article Classification with Manual Labeling and Machine Learning}
	\textbf{Input:} $\textit{cleanTrText, trainDF.class, cleanTsText, testDF.class}$ \\
	\textbf{Output:} $\textit{Evaluation of news article classification}$
	\begin{algorithmic}[1]
		\State $\textit{train} \gets \textit{cleanTrText, trainDF.class}$
		\State $\textit{test} \gets \textit{cleanTsText, testDF.class}$
		\State $\textit{xtrain, ytrain, xtest, ytest} \gets \textit{bow}(\textit{train, test, maxfeature=300})$
		
		\State $\textit{xtrain, ytrain, xtest, ytest} \gets \textit{tfidf}(\textit{train, test, maxfeature=300})$
		
		\State $\textit{d2vmodel,  traintag, testtag} \gets \textit{Doc2VecModel}(\textit{train, test, maxfeature=300})$
		\State $\textit{xtrain, ytrain, xtest, ytest} \gets \textit{Doc2Vec}(\textit{d2vmodel, traintag, testtag})$
		
		\State $\textit{evaluation} \gets \textit{mlAlgorithms}(\textit{xtrain, ytrain, xtest, ytest})$
	\end{algorithmic}
	\label{algo.manual_ML}
\end{algorithm}

\subsection{Deep Learning Algorithms}\label{sec.DL}
With three-word embedding techniques such as Word2vec, Glove, and Fasttext, we applied DL algorithms such as CNN (Convolutional Neural Network), LSTM (Long Short Term Memory), biLSTM (Bi-Directional LSTM), and GRU (Gated recurrent units). We classified the news articles using the manually labelled dataset. Algorithm~\ref{algo.manual_DL} shows the process for news article classification using deep learning algorithms.

\begin{algorithm}[H]
	\caption{News Article Classification with Manual Labeling and Deep Learning}
	\textbf{Input:} $\textit{cleanTrText, cleanTsText, w2vmodel, glovemodel, fasttextmodel}$ \\
	\textbf{Output:} $\textit{Evaluation of news article classification}$
	\begin{algorithmic}[1]	
		\State $\textit{train} \gets \textit{cleanTrText, trainDF.class}$
		\State $\textit{test} \gets \textit{cleanTsText, testDF.class}$
		
		\State $\textit{xtrain, ytrain, xtest, ytest} \gets \textit{train, test}$ \Comment{apply tokenizer, pad\_sequences}
		
		\State $\textit{w2v} \gets \textit{embeddingMatrix}(\textit{w2vmodel, cleanTrText})$ 
		\State $\textit{fasttext} \gets \textit{embeddingMatrix}(\textit{fasttextmodel, cleanTrText})$ 
		\State $\textit{glove} \gets \textit{embeddingMatrix}(\textit{glovemodel, cleanTrText})$ 
		
		\State $\textit{evaluation} \gets \textit{dlAlgorithms}(\textit{embeddingMatrix, train, test})$
	\end{algorithmic}
	\label{algo.manual_DL}
\end{algorithm}

\begin{figure}[!htb]
	\centering
	\includegraphics[scale=0.13]{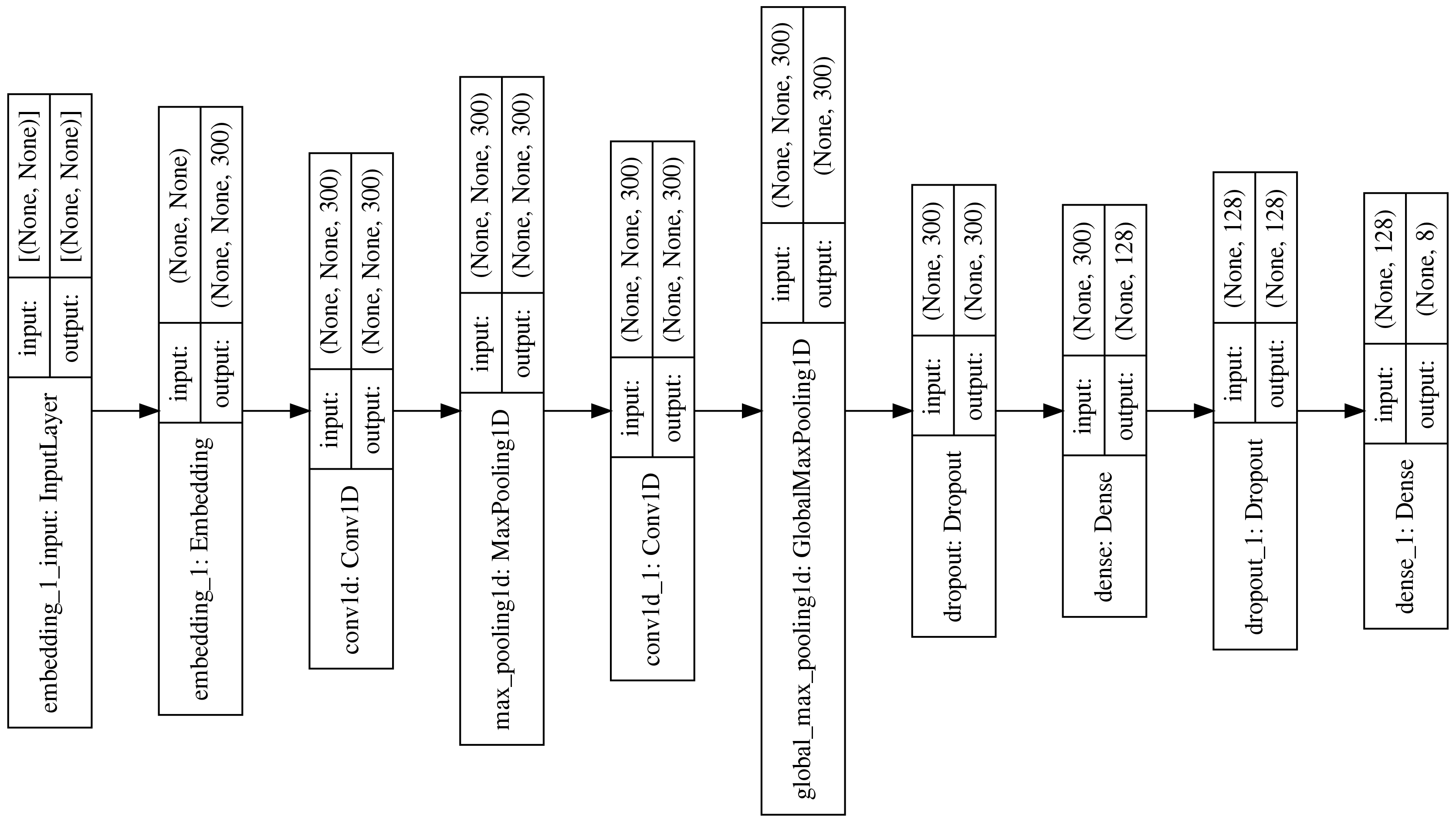}
	\captionsetup{justification=centering}
	\caption{CNN Architecture}
	\label{fig.CNN_architecture}
\end{figure}

\begin{figure}[!htb]
	\centering
	\includegraphics[scale=0.1]{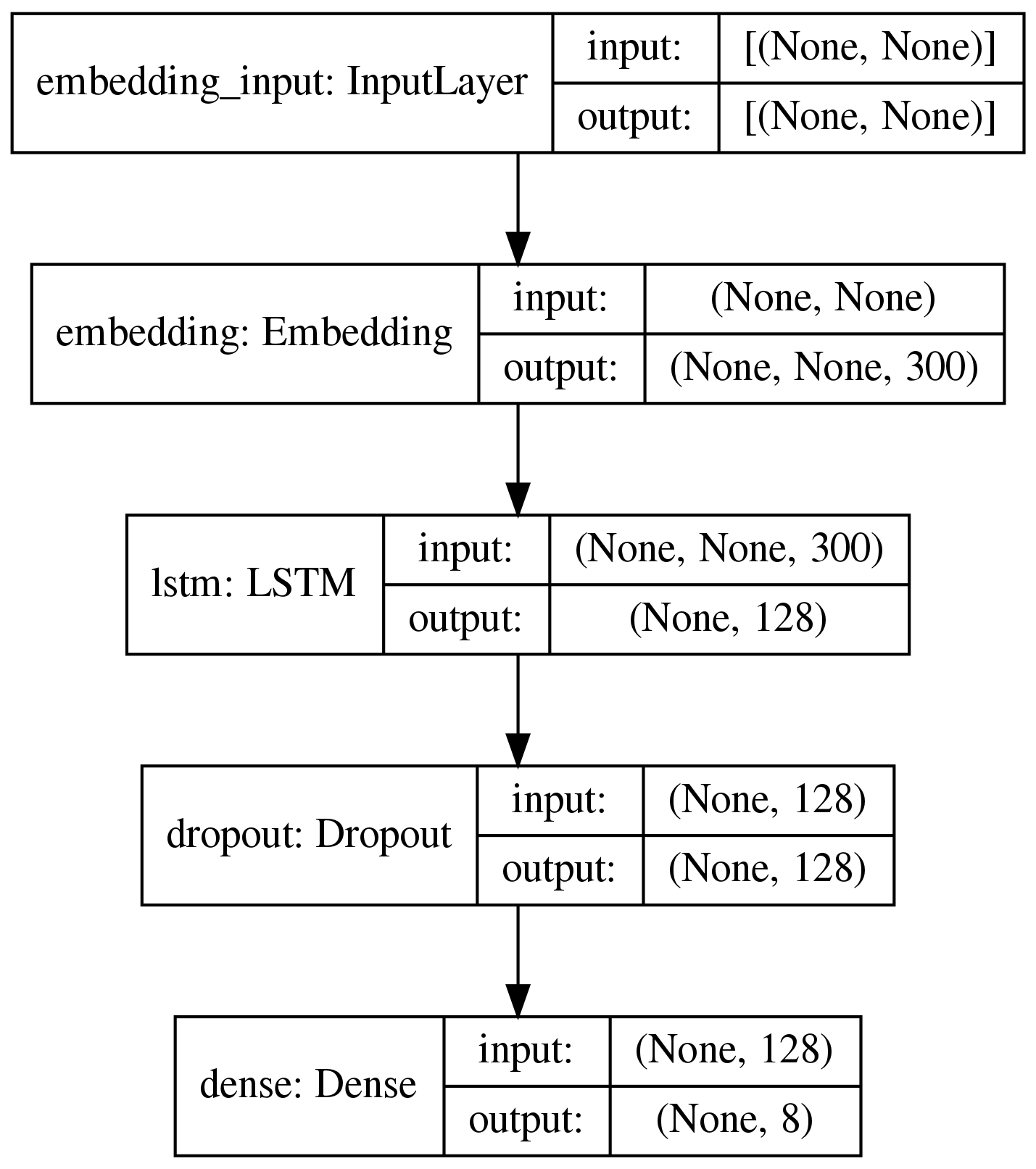}
	\includegraphics[scale=0.1]{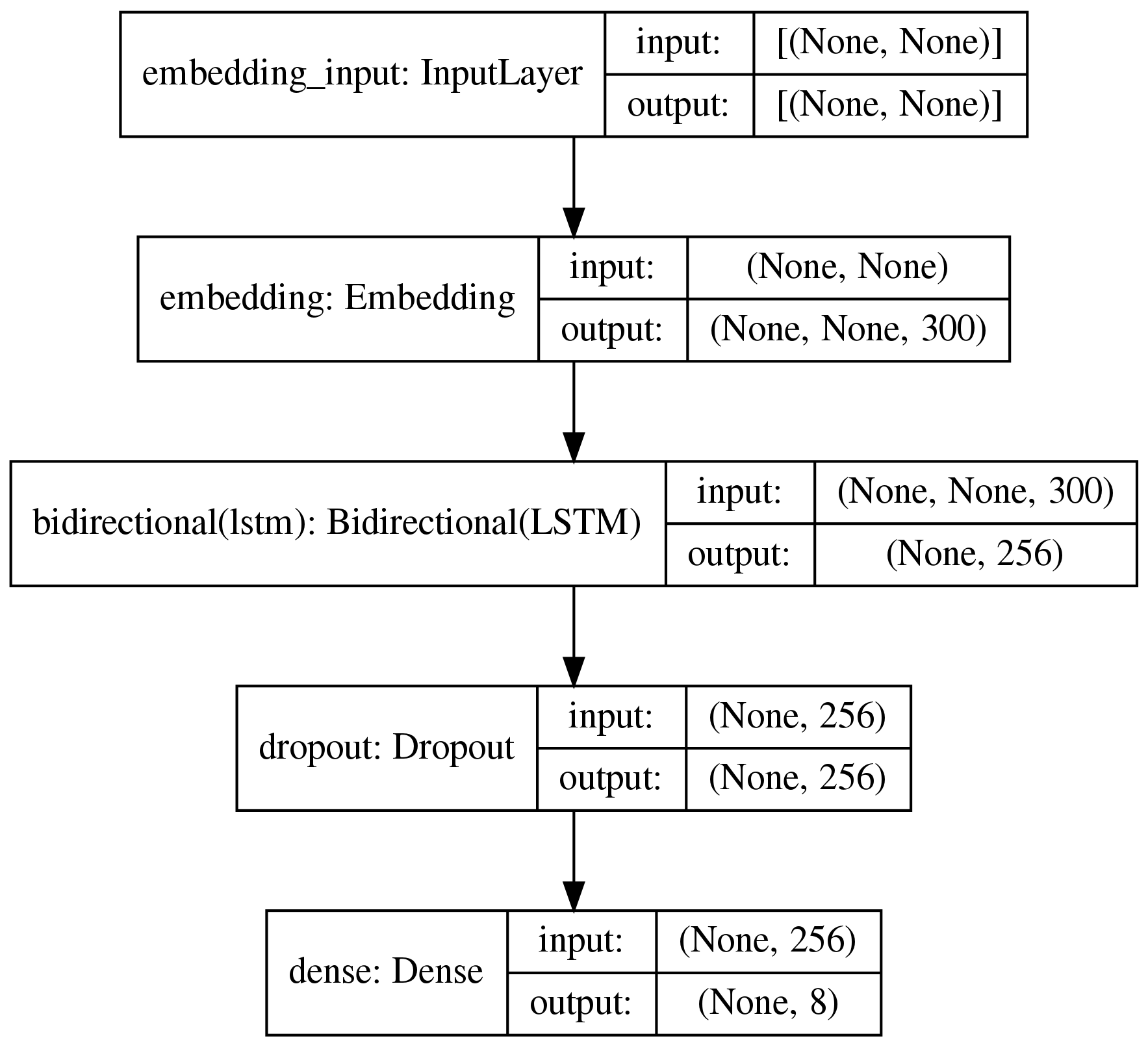}
	\includegraphics[scale=0.1]{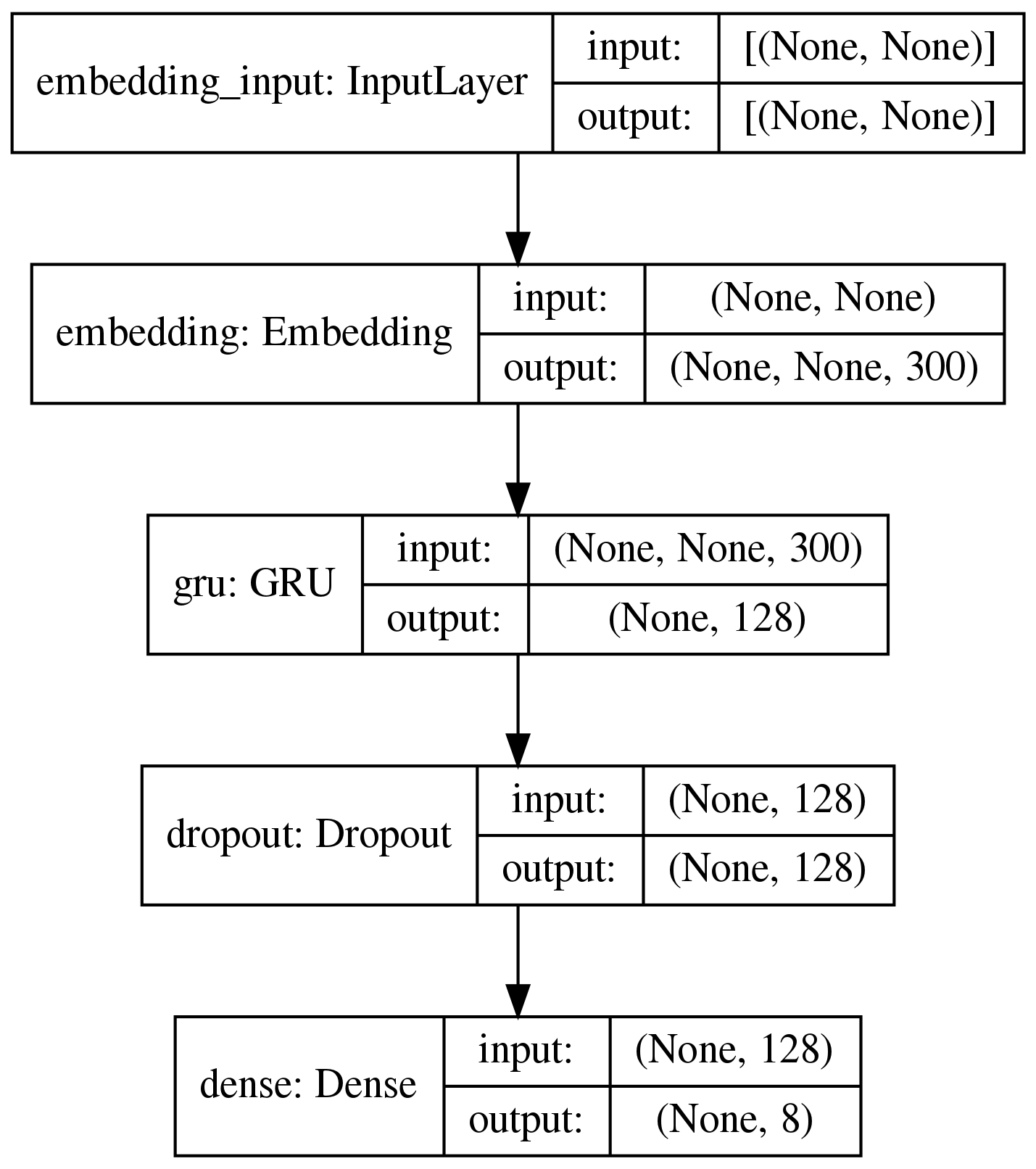}
	\captionsetup{justification=centering}
	\caption{Architecture of LSTM, BILSTM, and GRU}
	\label{fig.lstm_architecture}
\end{figure}

The CNN architectures is shown in Figure~\ref{fig.CNN_architecture}. For LSTM, BiLSTM, and GRU models, we used a similar architecture with a batch size of 64 and an epoch number of 8. The architecture of these three algorithms are shown in Figure~\ref{fig.lstm_architecture}. 

\subsection{Automatic Single labelled News Article Classification} \label{sec.automatic.single}
Topic modeling is an unsupervised machine learning method for analyzing and exploring latent information and expression patterns in multiple documents. In this study, we explore the behaviour of news article classification for the unlabelled dataset using the Latent Dirichlet Allocation (LDA) topic modeling technique.  Each news article is determined by the probability distribution over multiple topics or classes while a given topic or class is described as a probability distribution across words.

\begin{figure}[!htb]
	\centering
	\includegraphics[scale=0.9]{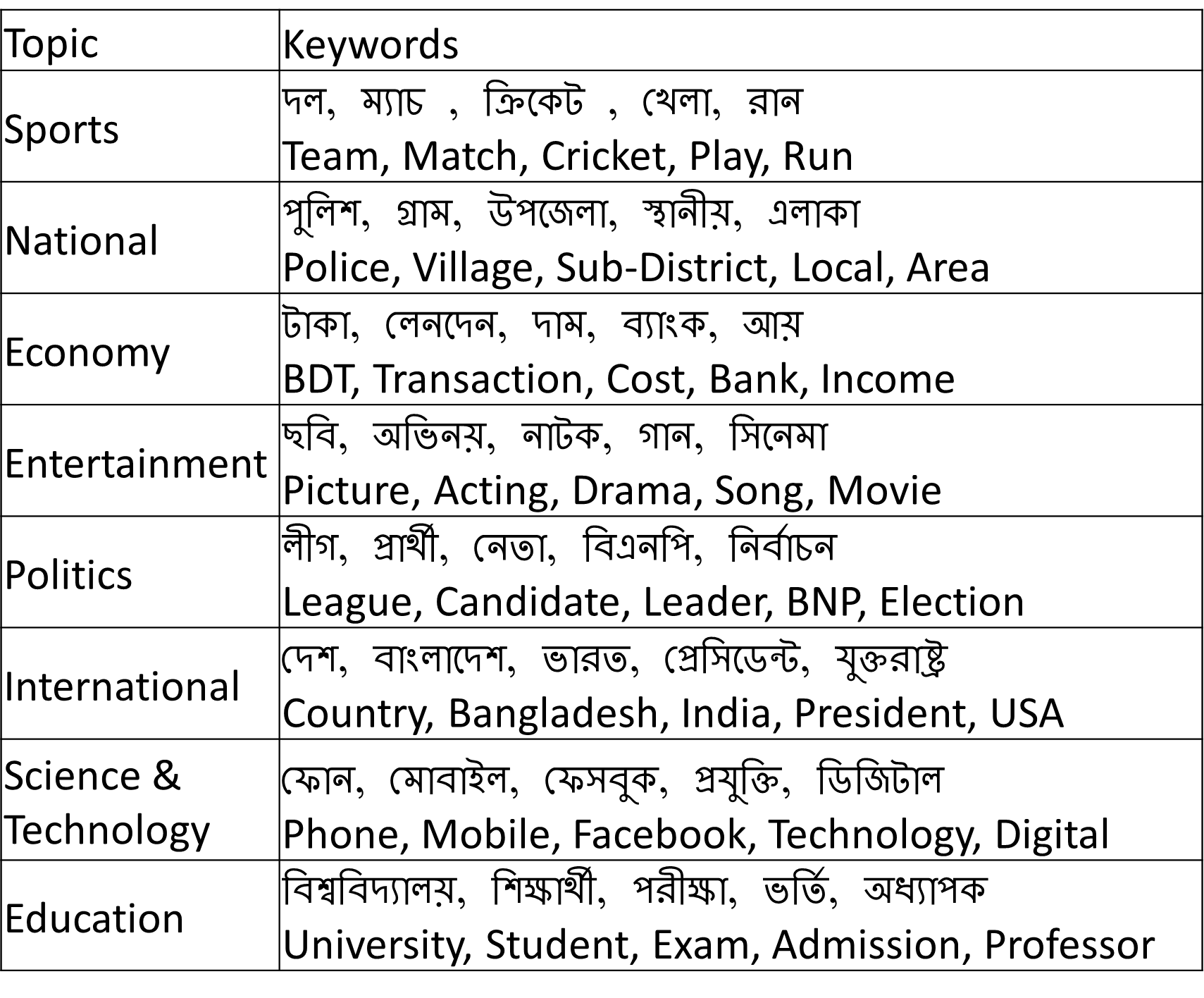}
	\captionsetup{justification=centering}
	\caption{Top 5 Keywords of Each Cluster using LDA}
	\label{tab.top_keywords}
\end{figure}

Algorithm~\ref{algo.auto_label.single} describes pseudo code to create an automatically labelled dataset. In our case, the number of classes is 8. We merge training and testing datasets and preprocess the merged articles. Doc2bow was used for the feature extraction, which includes the news article id and its frequency in every article. We use various parameter tunings, including n-gram, chunk size, passes, and iterations, to find the best LDA model. Based on the keywords and perplexity value, we chose the finest LDA model. After finding the best model, we have 8 lists of keywords based on the trained model, and each list defines a class. We manually labelled the names of the lists as news article class names based on the keywords. Table~\ref{tab.top_keywords} demonstrates the top 5 keywords for each of the classes. We retrieve an array of the probability distribution for each news article, which reflects how much the news article belongs to each class. To find the dominant class for each news article, we choose the class with the highest contribution probability. Finally, we create an automatically labelled dataset that contains all 8 class probabilities and the most dominant class, the manually labelled class, and the preprocessed article.

\begin{algorithm}[H]
	\caption{News Article Classification with Automatic Labeling for Single Label}
	\textbf{Input:} $\textit{trainDF, testDF, [ngram], [chunksize], [passes], [iterations], [th]}$ \\
	\textbf{Output:} $\textit{Evaluation of single label news article classification}$
	\begin{algorithmic}[1]
		\State $\textit{cleanArticlesDF} \gets \textit{trainDF.append(testDF)}$
		\State $\textit{cleanArticles} \gets \textit{preprocessing}(\textit{cleanArticlesDF})$
		
		\State $\textit{ngram} \gets \textit{make\_ngram}(\textit{cleanArticles, [ngram]})$
		
		\State $\textit{corpus} \gets \textit{doc2bow}(\textit{ngram})$
		
		\State $\textit{ldaModel} \gets \textit{getBestModel}(\textit{corpus, [chunksize], [passes], [iterations]})$
		
		\State $\textit{domClass} \gets \textit{getMostDominantClass}(\textit{ldaModel})$
		
		\State $\textit{classProb} \gets \textit{getEachClassProbPerArticle}(\textit{ldaModel})$
		
		\State $\textit{autoLabelDF} \gets \textit{getAutoLabelDF}(\textit{classProb, domClass, cleanArticlesDF.class, ngram})$
		
		\State $\textit{evaluation} \gets \textit{autoLabelDF.class, autoLabelDF.domClass}$
		
		\State $\textit{xtrain, ytrain, xtest, ytest} \gets \textit{getTrainTestSet}(\textit{autoLabelDF, [th]})$
		
		\State $\textit{evaluation} \gets \textit{mlAlgorithms}(\textit{xtrain, ytrain, xtest, ytest})$
		
	\end{algorithmic}
	\label{algo.auto_label.single}
\end{algorithm}

\begin{figure}[!htb]
		\centering
		\includegraphics[scale=0.4]{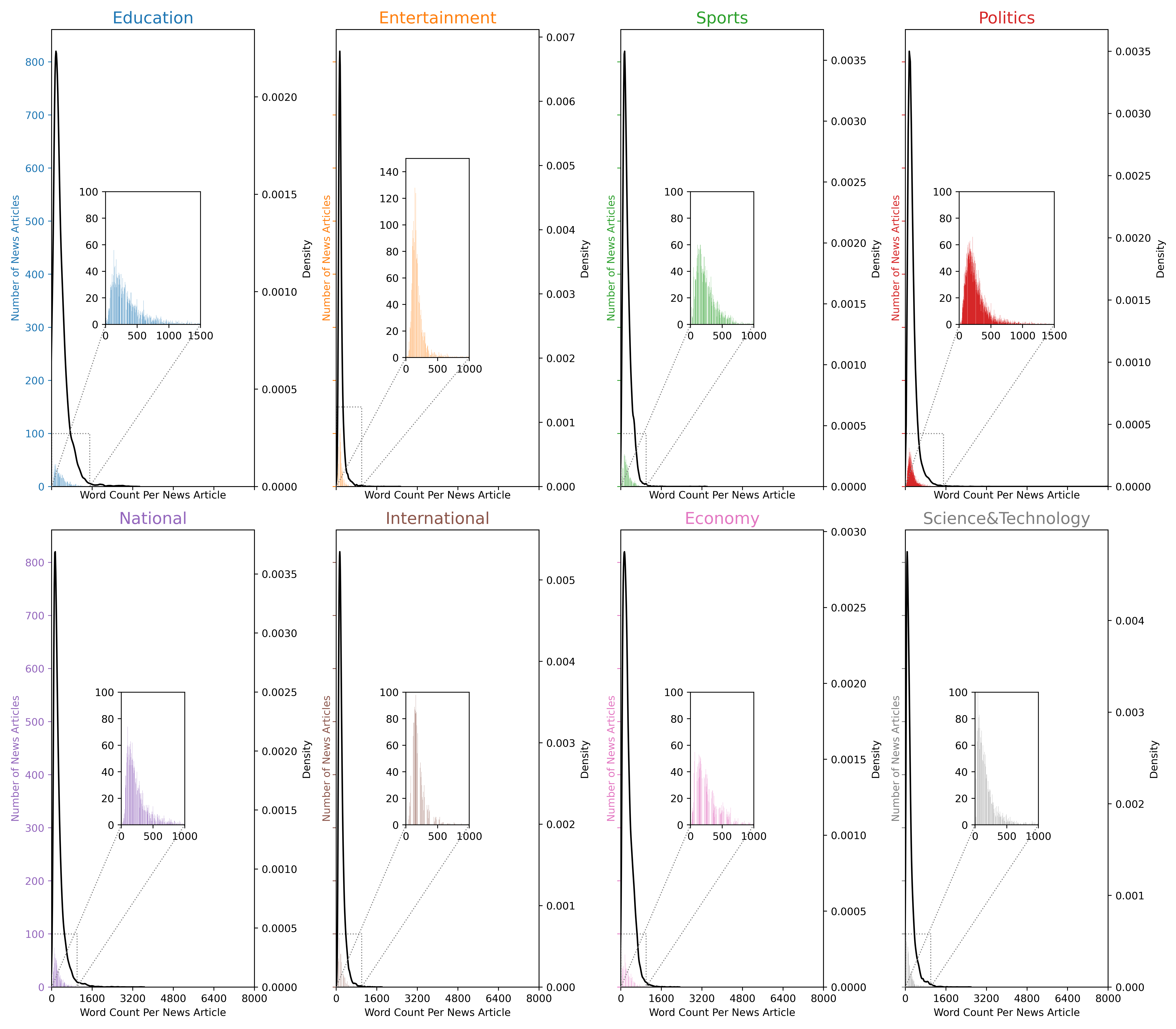}
		\captionsetup{justification=centering}
		\caption{Distribution of Document Word Counts (Original Topic)}
		\label{fig.Distribution_of_Document_Word_Counts_by_real_Topic}
\end{figure}

\begin{figure}[!htb]
		\centering
		\includegraphics[scale=0.39]{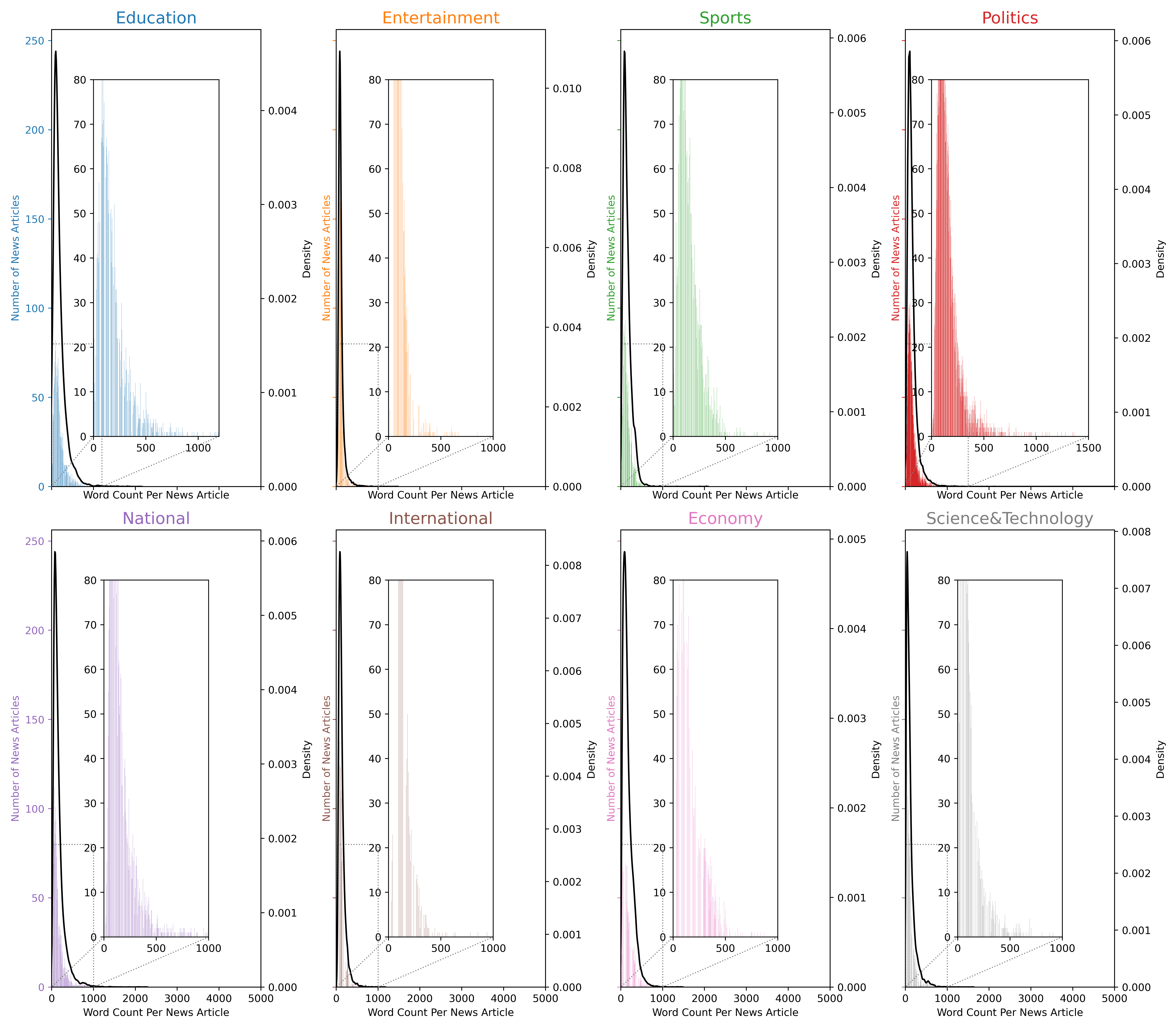}
		\captionsetup{justification=centering}
		\caption{Distribution of Document Word Counts (Dominant Topic)}
		\label{fig.Distribution_of_Document_Word_Counts_by_Dominant_Topic}
\end{figure}

\begin{figure}[H]
		\centering
		\includegraphics[scale=0.39]{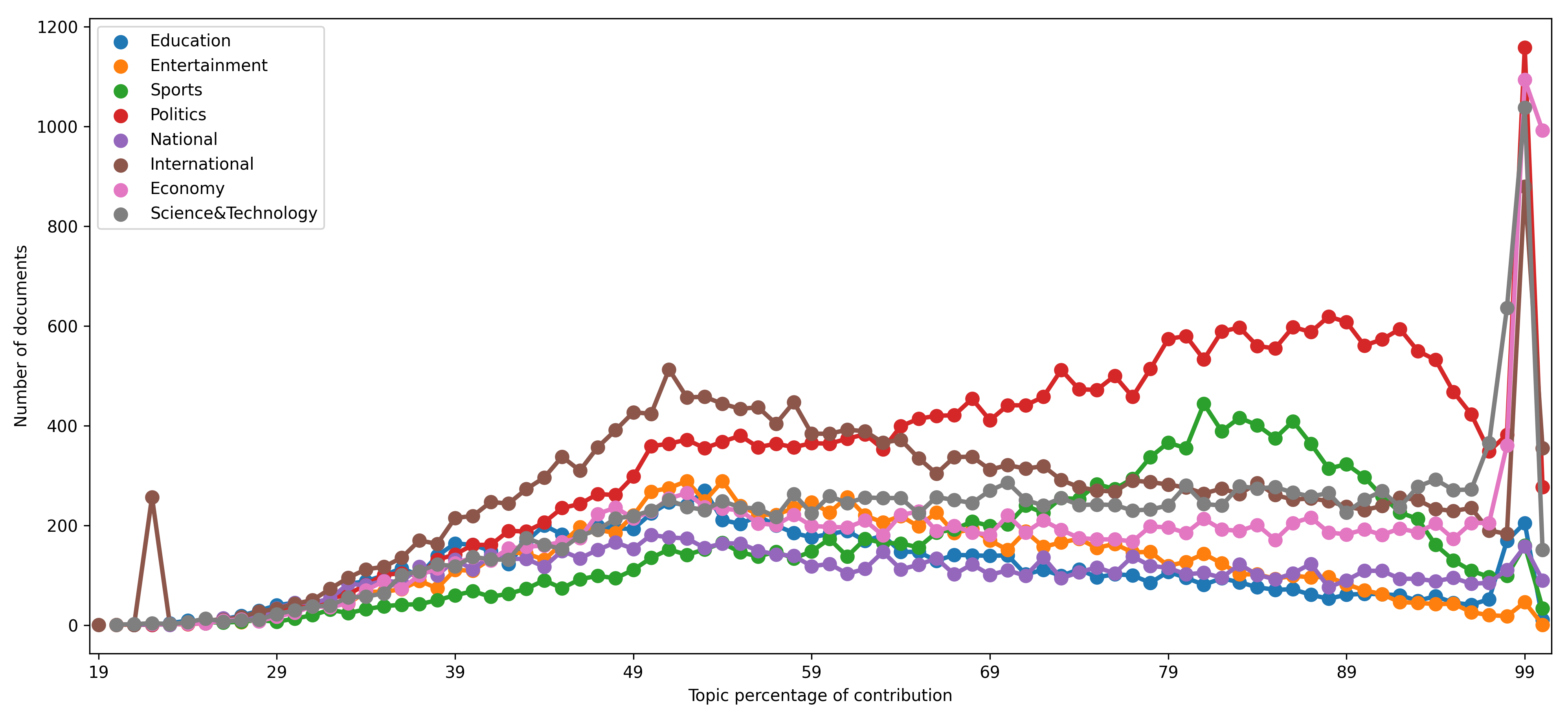}
		\captionsetup{justification=centering}
		\caption{Probability Map}
		\label{fig.probability_map}
\end{figure}

Figures~\ref{fig.Distribution_of_Document_Word_Counts_by_real_Topic} and~\ref{fig.Distribution_of_Document_Word_Counts_by_Dominant_Topic} describe the distribution of word counts by original topics and dominant topics respectively. Figure~\ref{fig.probability_map} shows the number of documents for each topic probability contribution. 

%
%
%
%
%
%
%
%

\subsection{Automatic Multi-labelled News Article Classification}\label{sec.automatic.multi}
A set of labels L, a set of examples X, where each example x is connected with a subset of the relevant L labels, can be used to represent the multi-label classification problem. The primary purpose of this is to create an L-dimensional target vector $y \epsilon \left\{0, 1\right\}L$, where $y_{i} = 1$ represents the relevant i-th label, whereas $y_{i} = 0$ represents the irrelevant label for the current case. To address the multi-label text classification issue, there are primarily three types of approaches: problem transformation methods (Label Powerset, Binary Relevance, and Classifier Chains), Algorithm adaptation methods (KNN, decision trees), and neural network models.
In this paper, we perform binary relevance learning that constructs L binary classifiers by training on the L labels individually and combining all the classifiers results into a multi-label prediction by overlooking the associations between labels.

In the previous section, we created an automatically labelled dataset containing the 8 class probabilities for each item. Algorithm~\ref{algo.auto_label.multi} describes the pseudo code for creating a multi-labelled dataset using this dataset. We use MultiLabelBinarizer to generate the multi-label classes. We apply the same ML techniques with the Doc2Vec feature extraction methodology to train the model. As the prediction for an example or instance consists of a collection of labels that may be entirely correct, partially correct (with varying degrees of accuracy), or completely incorrect, evaluating a multi-label classifier is more difficult than evaluating a single-label classifier. We will use example-based matrices to evaluate the multi-label classifiers performance, such as subset f1-score, precision, recall, accuracy, and hamming loss.

\begin{algorithm}[H]
	\caption{News Article Classification with Automatic Labeling for Multi-Label}
	\textbf{Input:} $\textit{autoLabelDF, cleanArticles, [th]}$ \\
	\textbf{Output:} $\textit{Evaluation of multi-label news article classification}$
	\begin{algorithmic}[1]
		\State $\textit{mulLabelDF} \gets \textit{getMultiLabelBinarizer}(\textit{autoLabelDF.classProb, [th]})$
		\State $\textit{manMulLblClass} \gets \textit{getMultiLabelBinarizer}(\textit{autoLabelDF.class})$
		\State $\textit{evaluation} \gets \textit{manMulLblClass, mulLabelDF}$
		
		\State $\textit{xtrain, ytrain, xtest, ytest} \gets \textit{mlAlgo\_d2v}(\textit{autoLabelDF.ngram, mulLabelDF})$
		
		\State $\textit{evaluation} \gets \textit{mlAlgorithms}(\textit{xtrain, ytrain, xtest, ytest})$
	\end{algorithmic}
	\label{algo.auto_label.multi}
\end{algorithm}

\begin{figure}[!htb]
		\centering
		\includegraphics[scale=0.99]{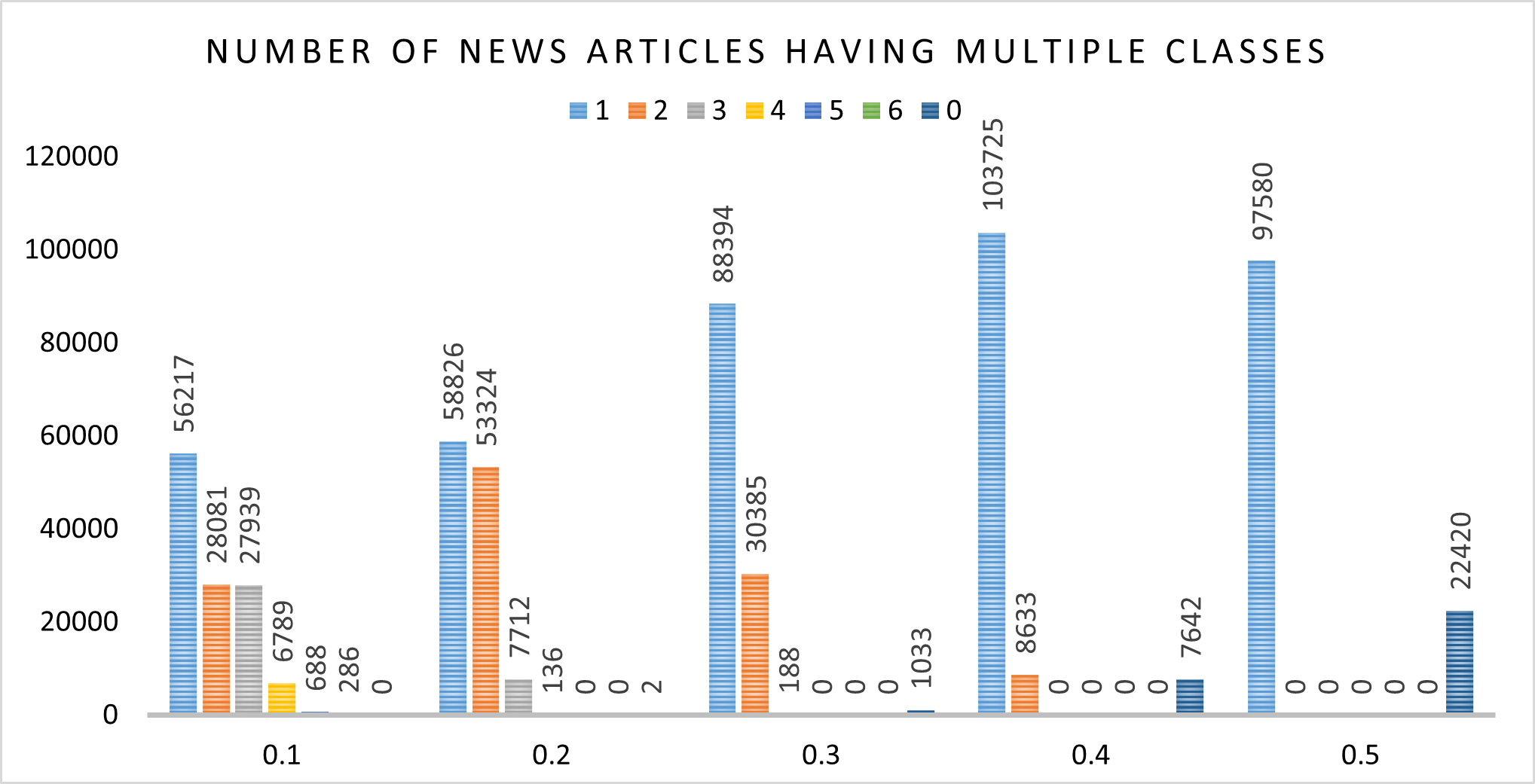}
		\captionsetup{justification=centering}
		\caption{Number of News Articles Having Multiple Classes}
		\label{fig.multilabel_nm}
\end{figure}

Figure~\ref{fig.multilabel_nm} describes the number of news articles with multiple classes for different probability thresholds such as 0.1, 0.2, 0.3, 0.4, and 0.5. For a threshold of 0.1, 56,217 news articles have 1 class, 28,081 news articles have 2 classes, 27,939 news articles have 3 classes, 6,789 news articles have 4 classes, 688 news articles have 5 classes, 286 news articles have 6 classes and no article has no class. For a threshold of 0.5, most of the articles have 1 class and 22,420 articles have no class because the probability of each class is less than 0.5. To evaluate multi-label classification, we use the threshold of 0.3, where 88,394 articles have 1 class, 30,385 articles have 2 classes, 188 articles have 3 classes, and 1,033 articles have no class.

\begin{figure}[!htb]
		\centering
		\includegraphics[scale=0.47]{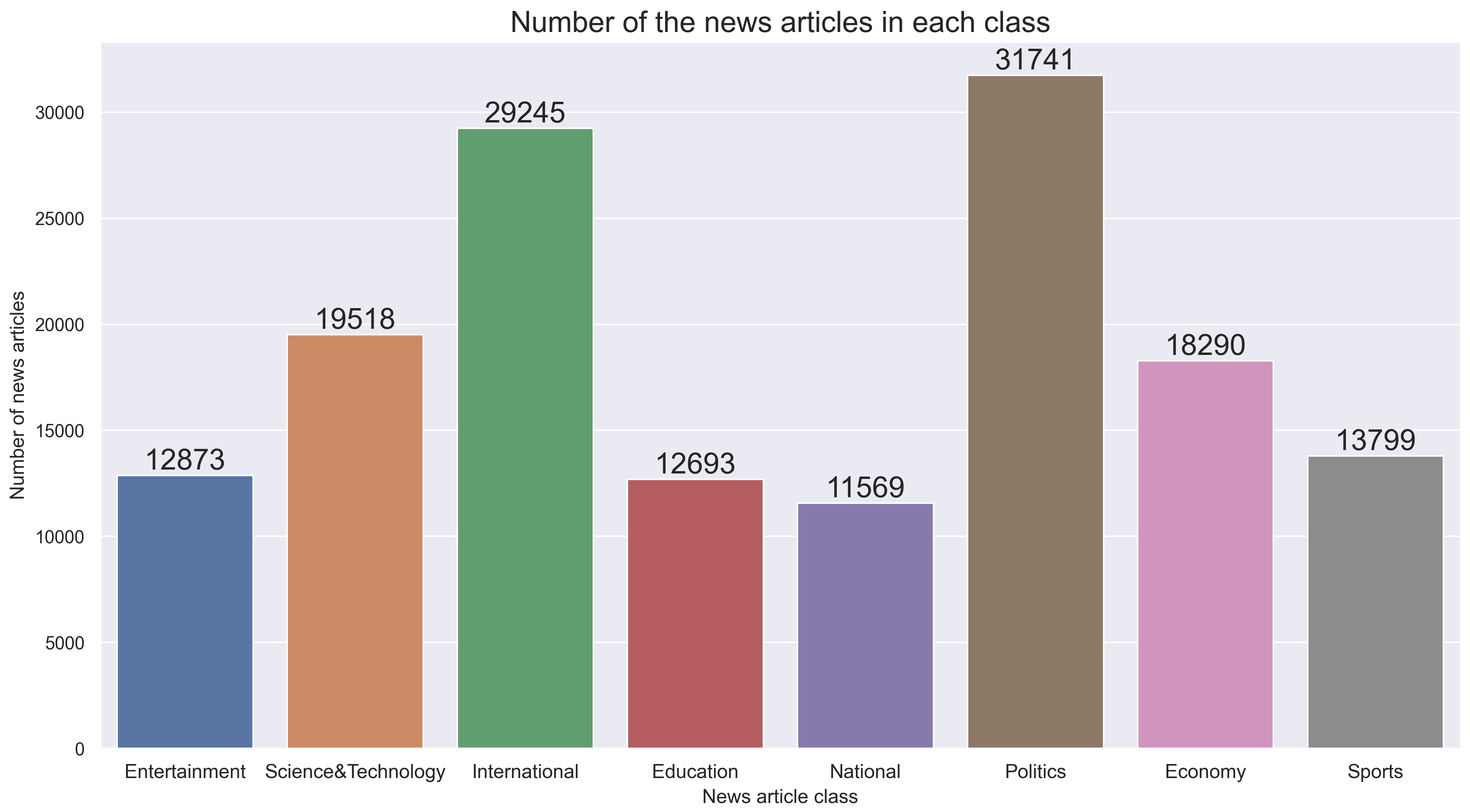}
		\captionsetup{justification=centering}
		\caption{Number of News Articles for Each Class}
		\label{fig.multilabel_ne}
\end{figure}

The number of news articles for each class is shown in Figure~\ref{fig.multilabel_ne}. The International and Politics classes have the most articles, with 29,245 and 31,741 respectively.

\section{Manually labelled News Article Classification}\label{sec.result.manual.ml}

This section analyses the performance of supervised machine and deep learning algorithms, which were discussed in Section~\ref{sec.ML} and~\ref{sec.DL}, to establish a detailed comparison of the news article classification models.

\begin{figure}[H]
		\centering
		\includegraphics[scale=0.99]{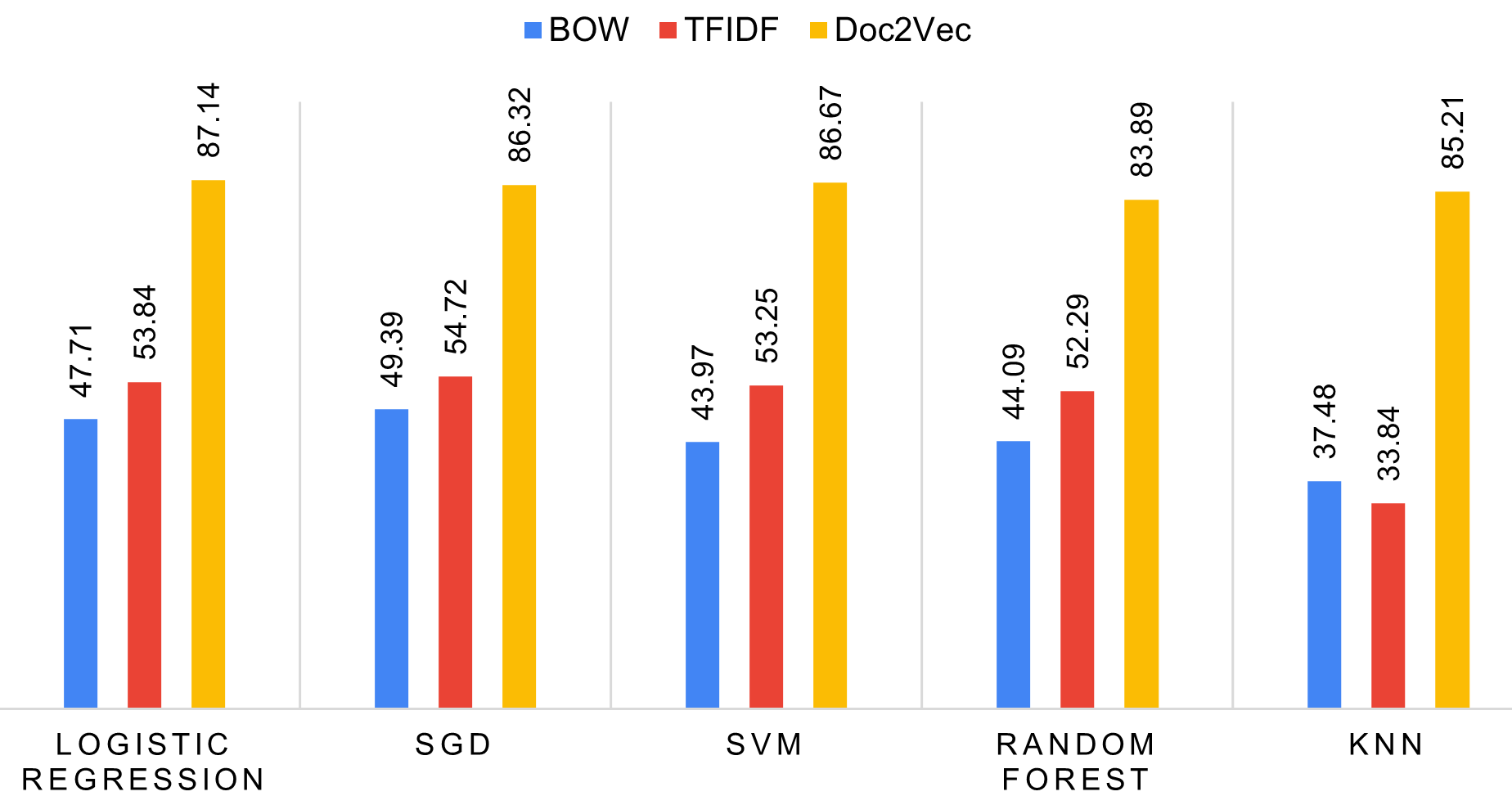}
		\captionsetup{justification=centering}
		\caption{Manually labelled Article Classification Accuracy for ML Algorithms}
		\label{fig.ML_accuracy}
\end{figure}

Figure~\ref{fig.ML_accuracy} shows the accuracy in percentage of the news article classification for machine learning algorithms using embedding techniques, i.e., BOW, TFIDF, and Doc2Vec. We used a maximum of 300 feature vectors for each word embedding technique. The highest accuracy was achieved 87.14\% by logistic regression algorithm with the Doc2Vec technique. Moreover, the Doc2Vec technique obtained more than 80\% accuracy for all the ML algorithms, whereas the BOW and TFIDF techniques obtained very low accuracy. The Doc2Vec is a word2vec extension that works as a memory for what is lacking in the present context. On the other hand, BOW and TFIDF do not capture the context of the text. As a result, in large documents with extensive textual variation and complexity, the ML models are overfitted for BOW and TFIDF approaches. With the BOW and TFIDF techniques, we achieved the highest accuracy of 49.39\% and 54.72\% for SGD, respectively. In summary, Doc2Vec outperforms the BOW and TFIDF approaches in terms of accuracy.

\begin{figure}[H]
		\centering
		\includegraphics[scale=0.88]{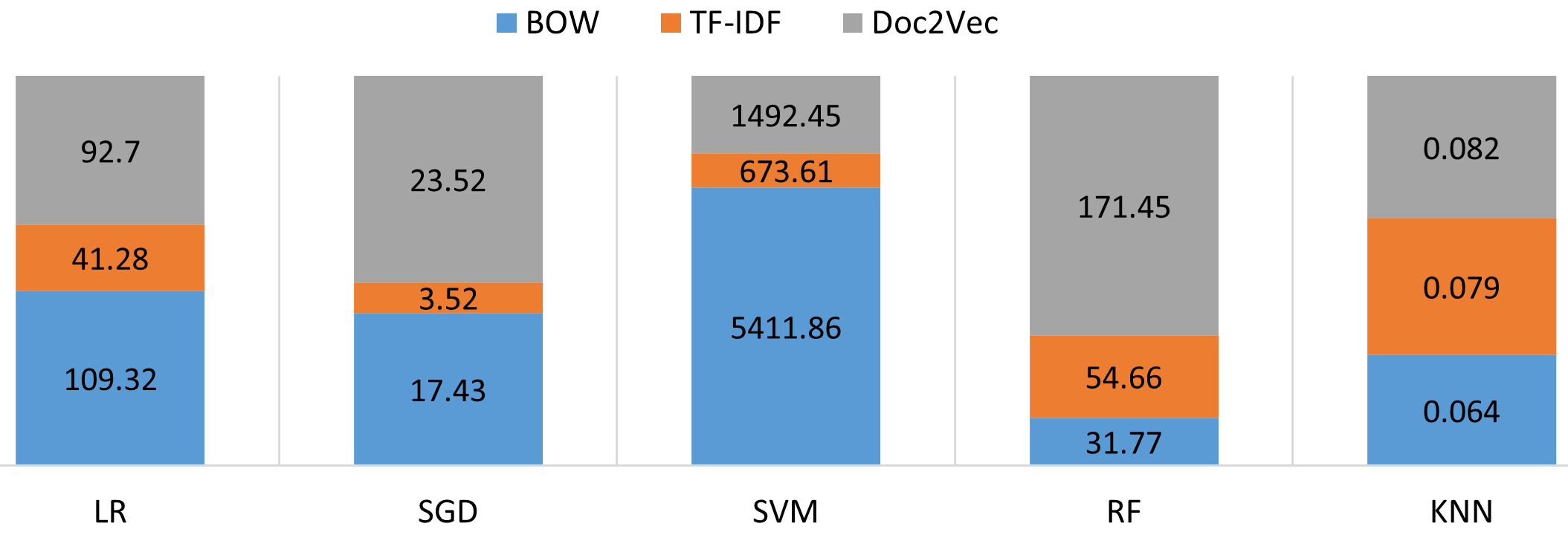}
		\captionsetup{justification=centering}
		\caption{Execution Time (seconds) for ML Algorithms}
		\label{fig.time_ML}
\end{figure}

Figure~\ref{fig.time_ML} depicts the execution time for ML algorithms where SVM and KNN have the highest and lowest execution times compared to other ML algorithms. For the SVM algorithm, BOW has the highest execution time of about 5411.86 seconds, whereas TFIDF gets less time. The execution time is high in SVM because of the kernel parameter which slows down the process.

\begin{figure}[H]
		\centering
		\includegraphics[scale=0.78]{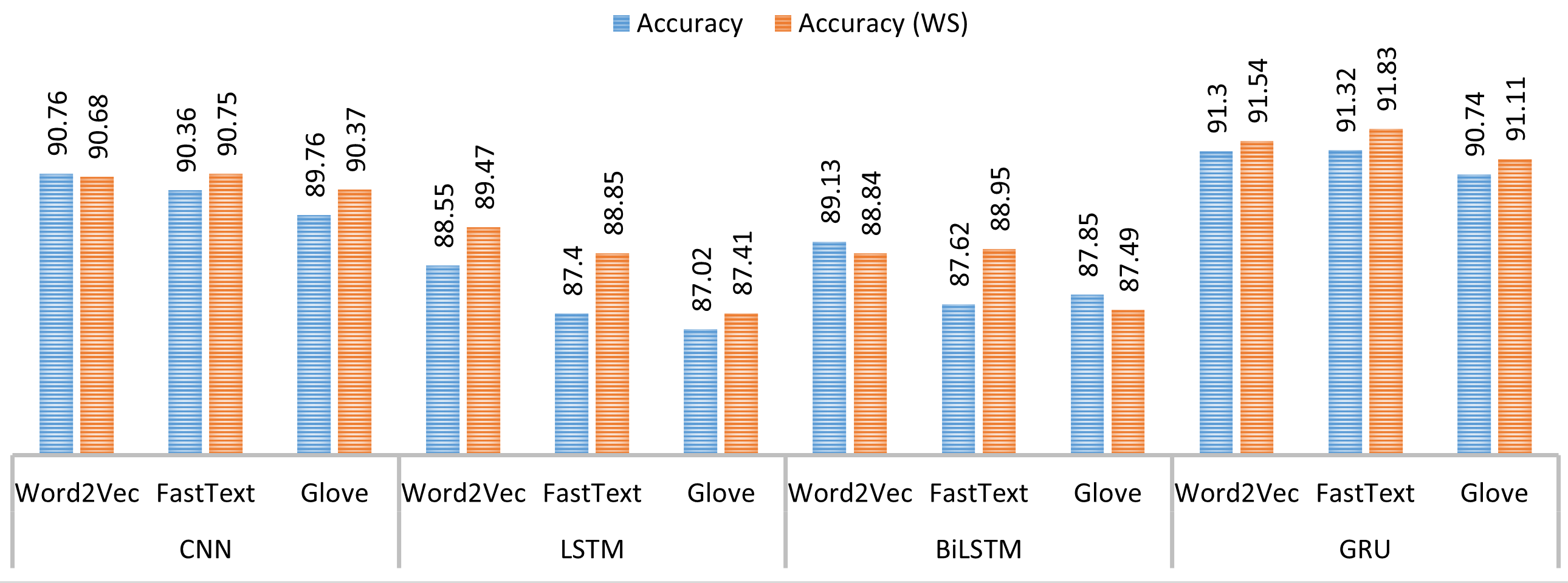}
		\captionsetup{justification=centering}
		\caption{Manually labelled Article Classification Accuracy for DL Algorithms}
		\label{fig.DL_accuracy}
\end{figure}

We used three word embedding techniques (Word2Vec, Fasttext, and Glove) to evaluate the performance of the DL (CNN, LSTM, BiLSTM, and GRU) models. Additionally, we analysed the performance of article classification models in two ways: with stemmer and without stemmer (WS). We noticed that the stemmer has no significant impact on DL accuracy. Figure~\ref{fig.DL_accuracy} demonstrates the accuracy of the DL algorithms for each word embedding technique. We obtained the highest accuracy of 91.83\% for GRU with Fasttext and stemmer. 

\begin{figure}[H]
		\centering
		\includegraphics[scale=0.78]{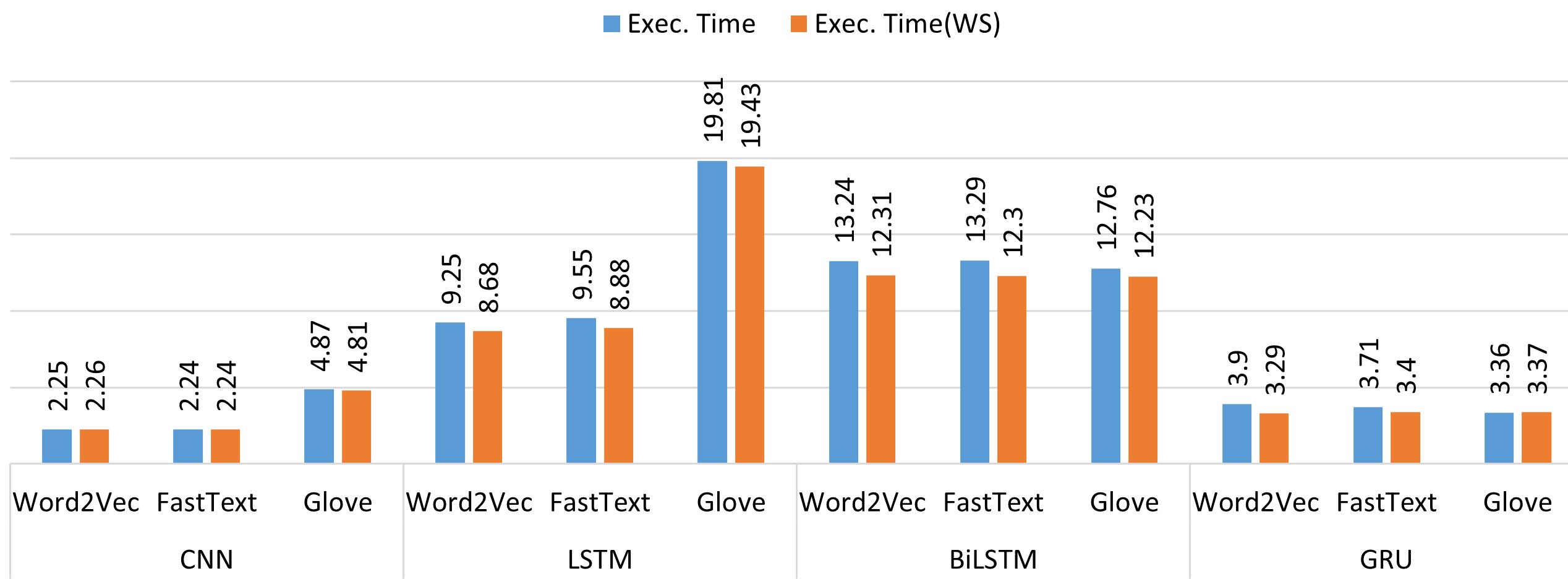}
		\captionsetup{justification=centering}
		\caption{Execution Time (hour) for DL Algorithms}
		\label{fig.time_DL}
\end{figure}

Figure~\ref{fig.time_DL} shows that the LSTM and BiLSTM algorithms take the most execution time, whereas CNN and GRU have significantly less execution time. LSTM has less execution time than BiLSTM, as in BiLSTM, the input sequence flow is both forward and backward, which captures the context from both the past and present sequence. For example, BiLSTM execution times for word2vec and FastText techniques are 13.24 and 13.29 hour, whereas LSTM execution times for word2vec and FastText techniques are 9.25 and 9.55 hour. However, with the Glove technique, LSTM has a longer execution time than BiLSTM.

\begin{figure}[H]
		\centering
		\includegraphics[scale=0.40]{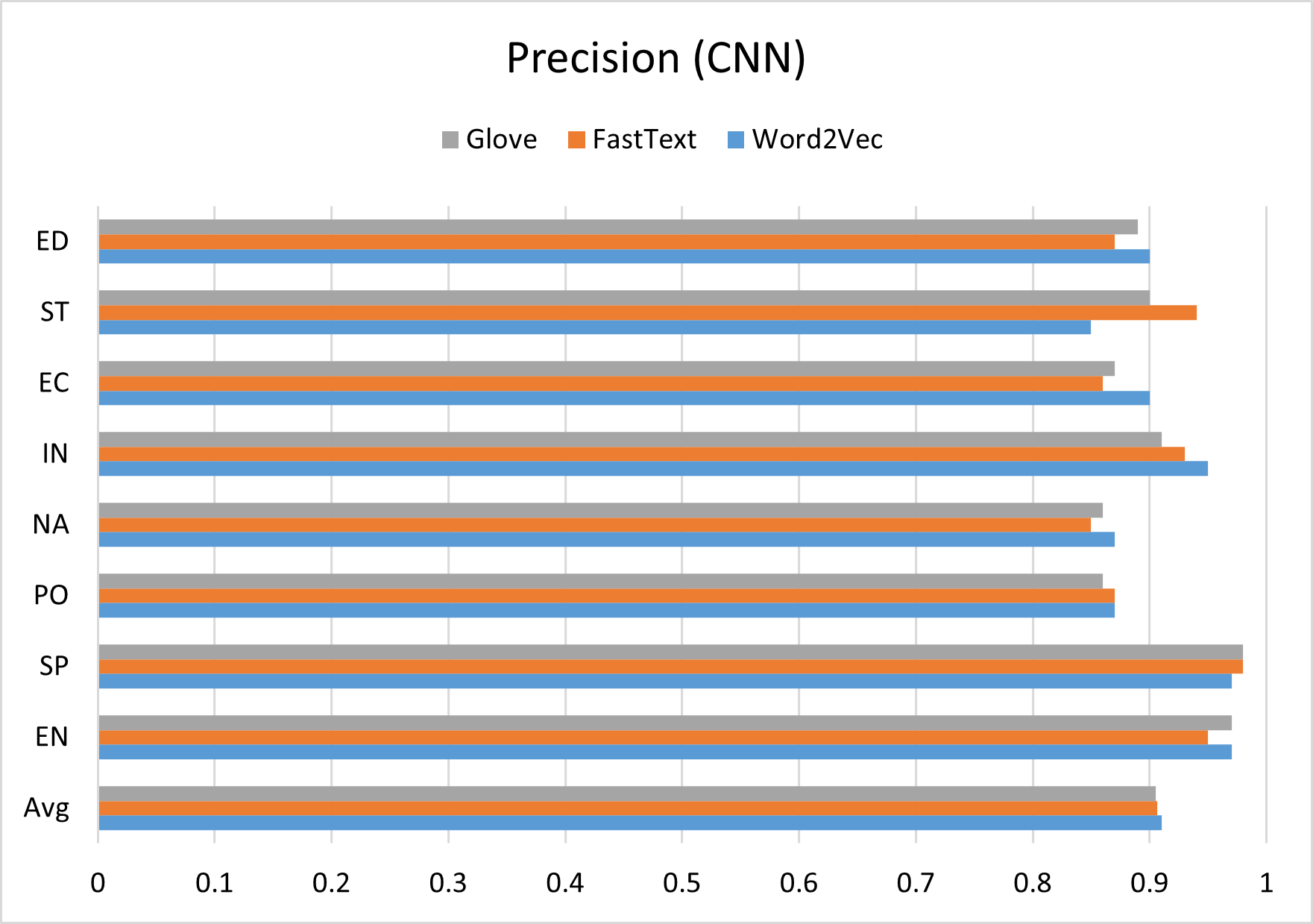}		
		\includegraphics[scale=0.40]{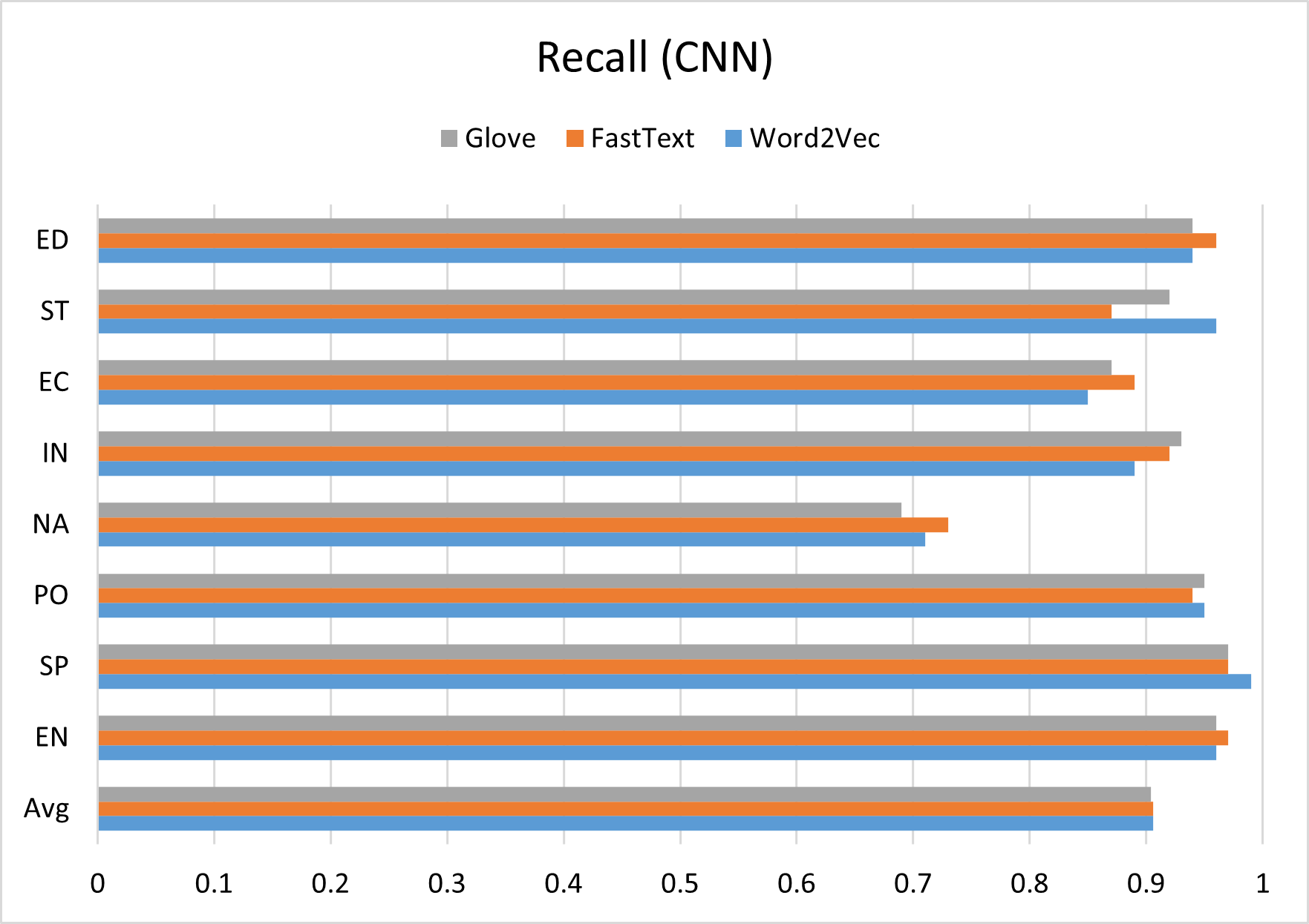}
		\includegraphics[scale=0.40]{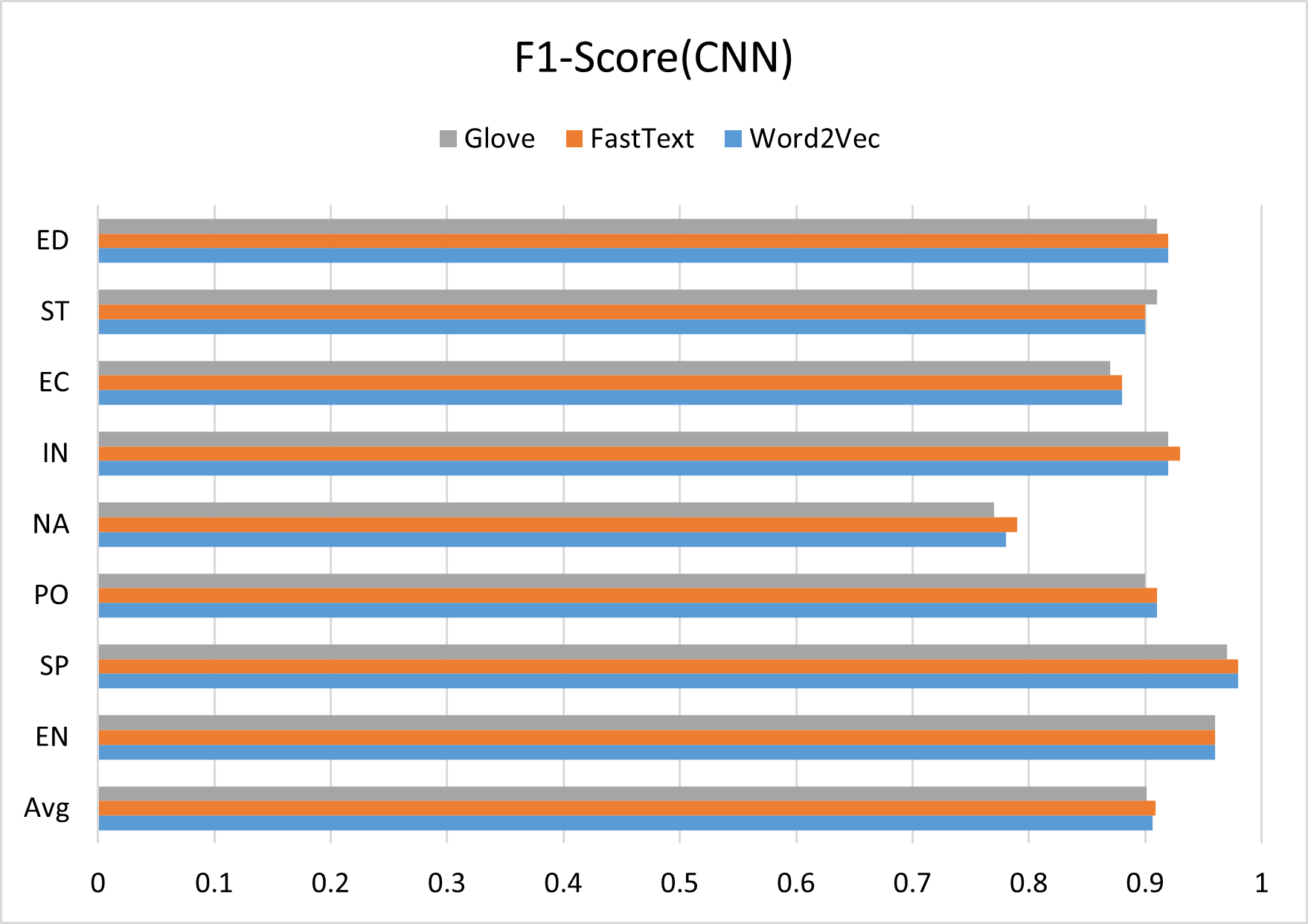}
		\captionsetup{justification=centering}
		\caption{Evaluation Metrics for Each Article Class using CNN}
		\label{fig:evaluationMetrics_CNN}
\end{figure}

\begin{figure}[H]
		\centering
		\includegraphics[scale=0.40]{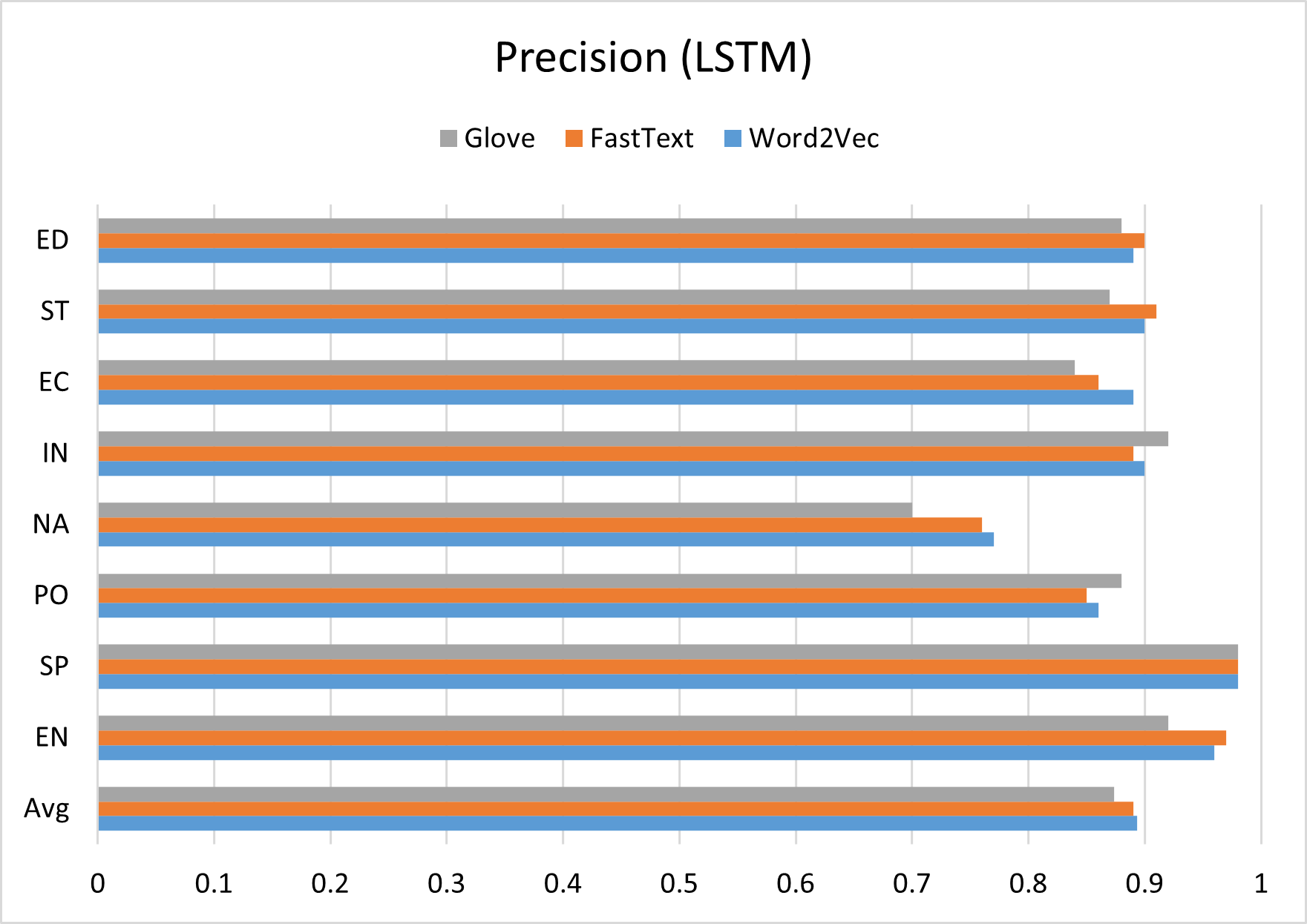}		
		\includegraphics[scale=0.40]{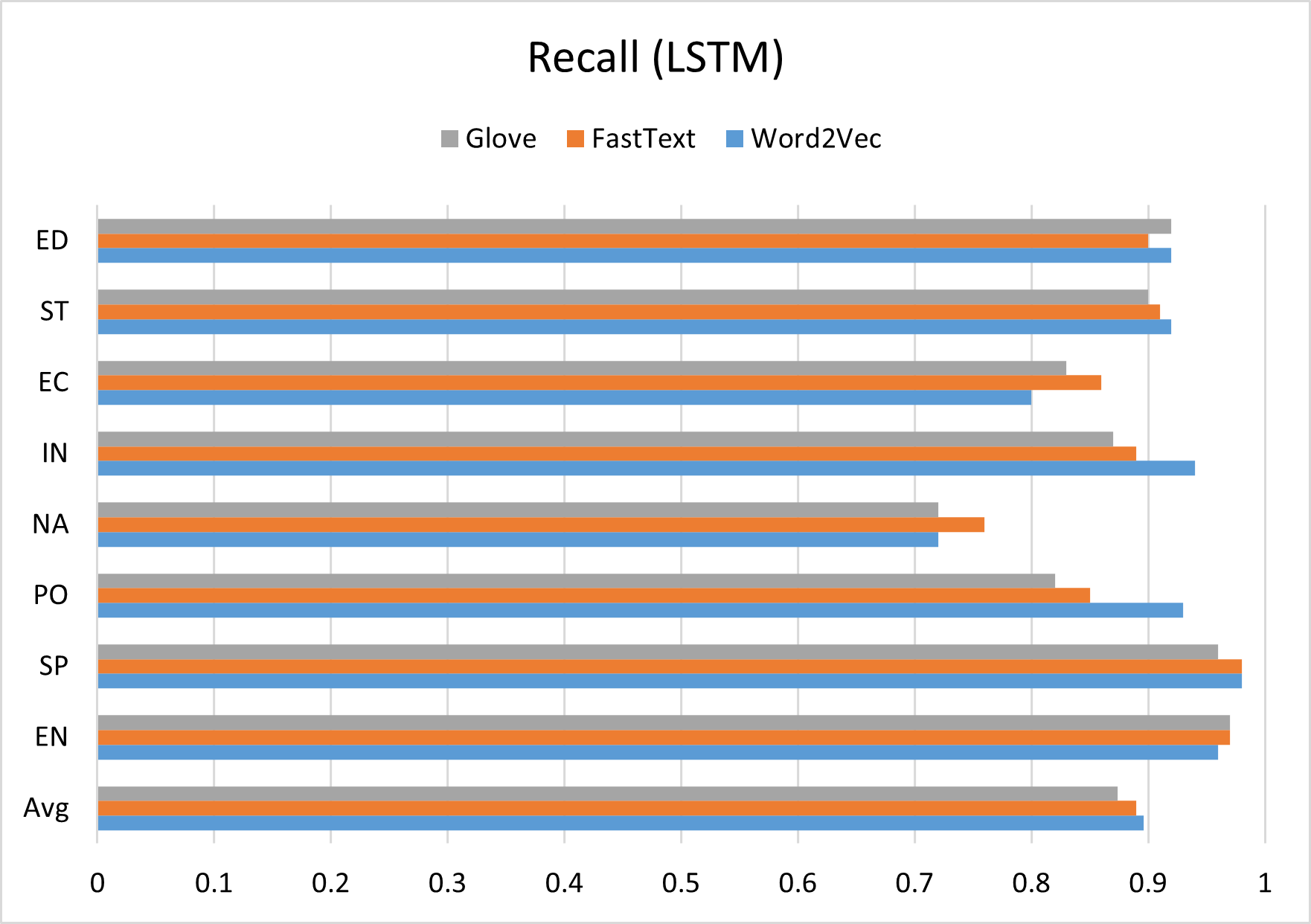}
		\includegraphics[scale=0.40]{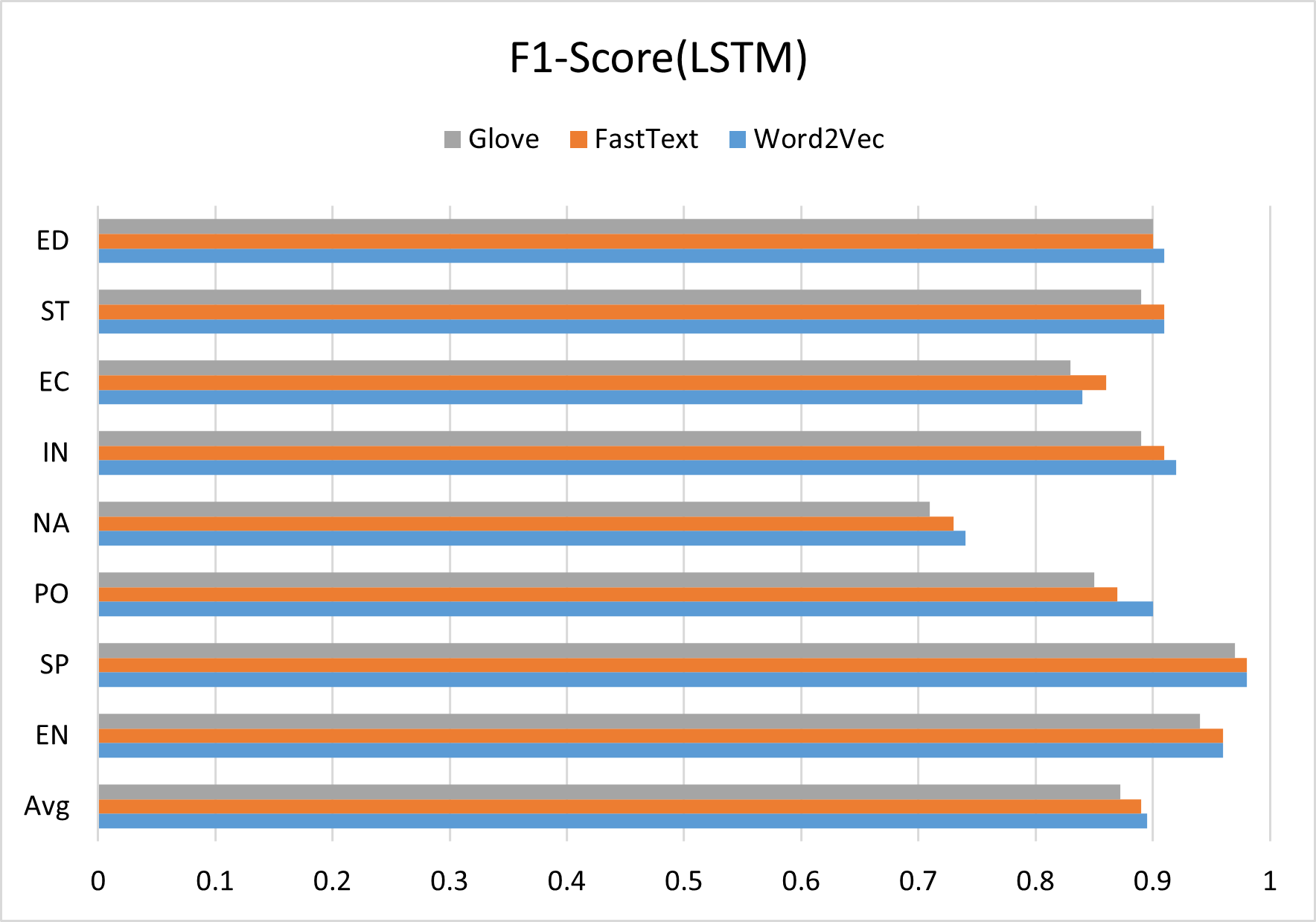}
		\captionsetup{justification=centering}
		\caption{Evaluation Metrics for Each Article Class using LSTM}
		\label{fig:evaluationMetrics_LSTM}
\end{figure}

\begin{figure}[h]
		\centering
		\includegraphics[scale=0.40]{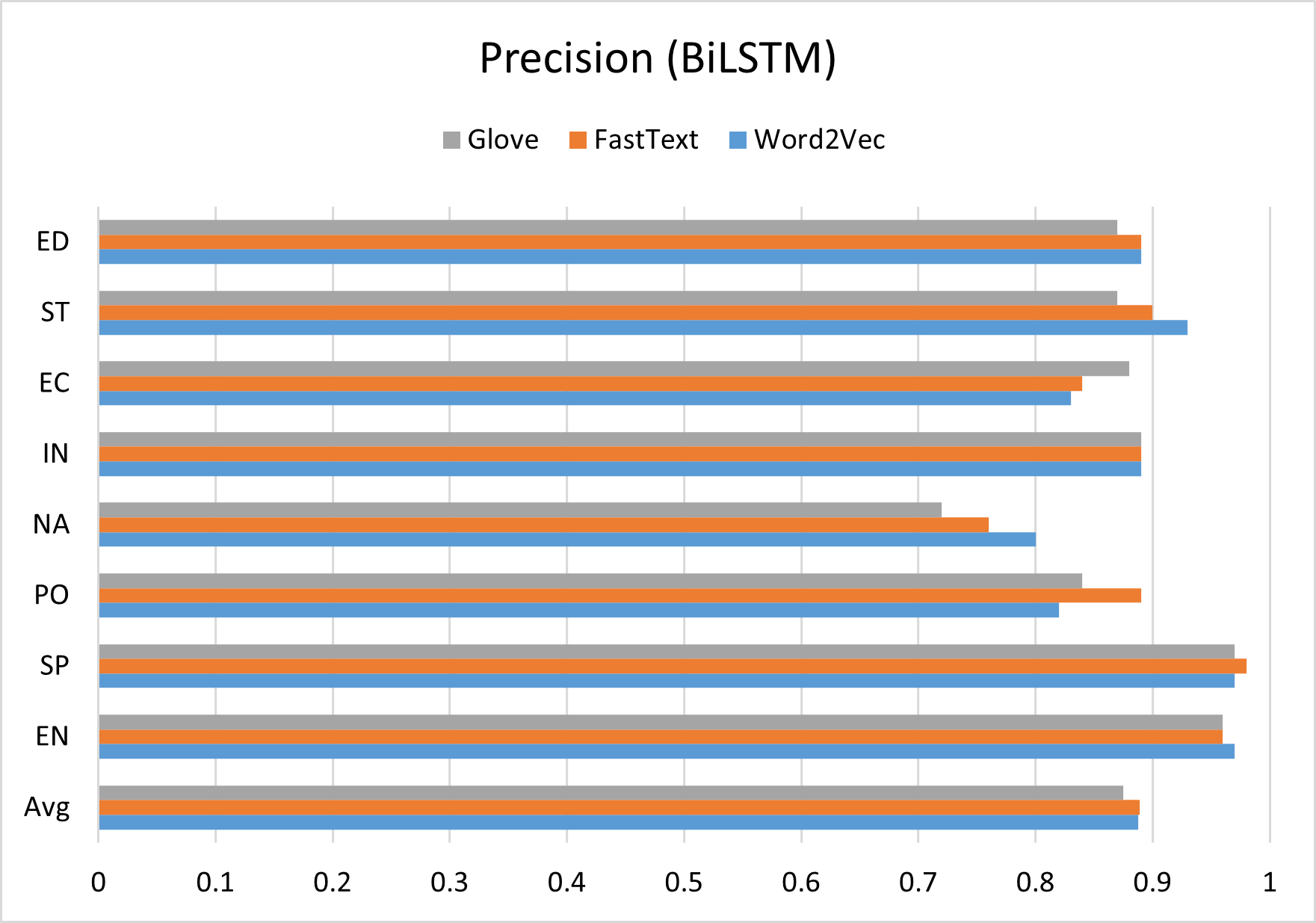}		
		\includegraphics[scale=0.40]{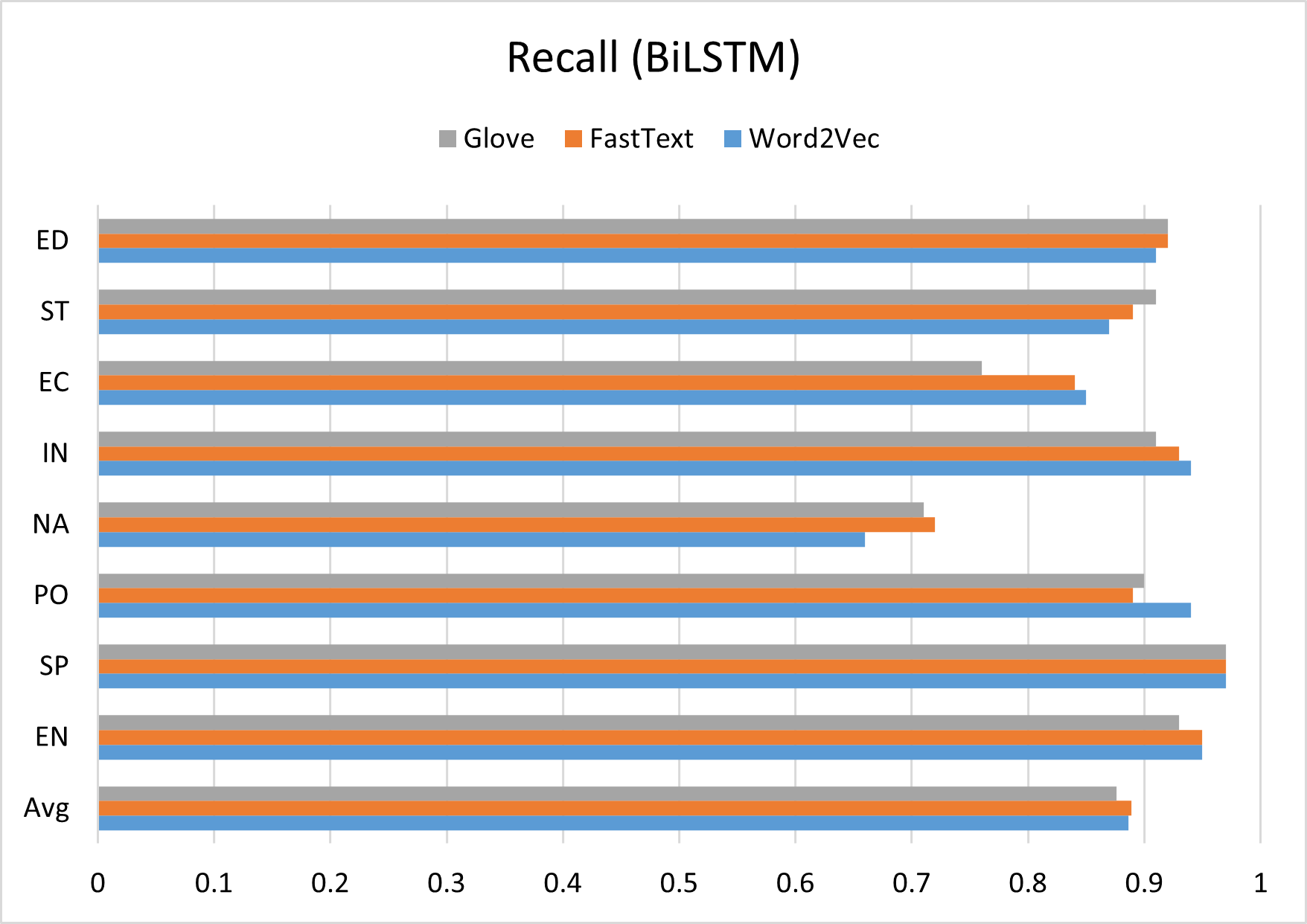}
		\includegraphics[scale=0.40]{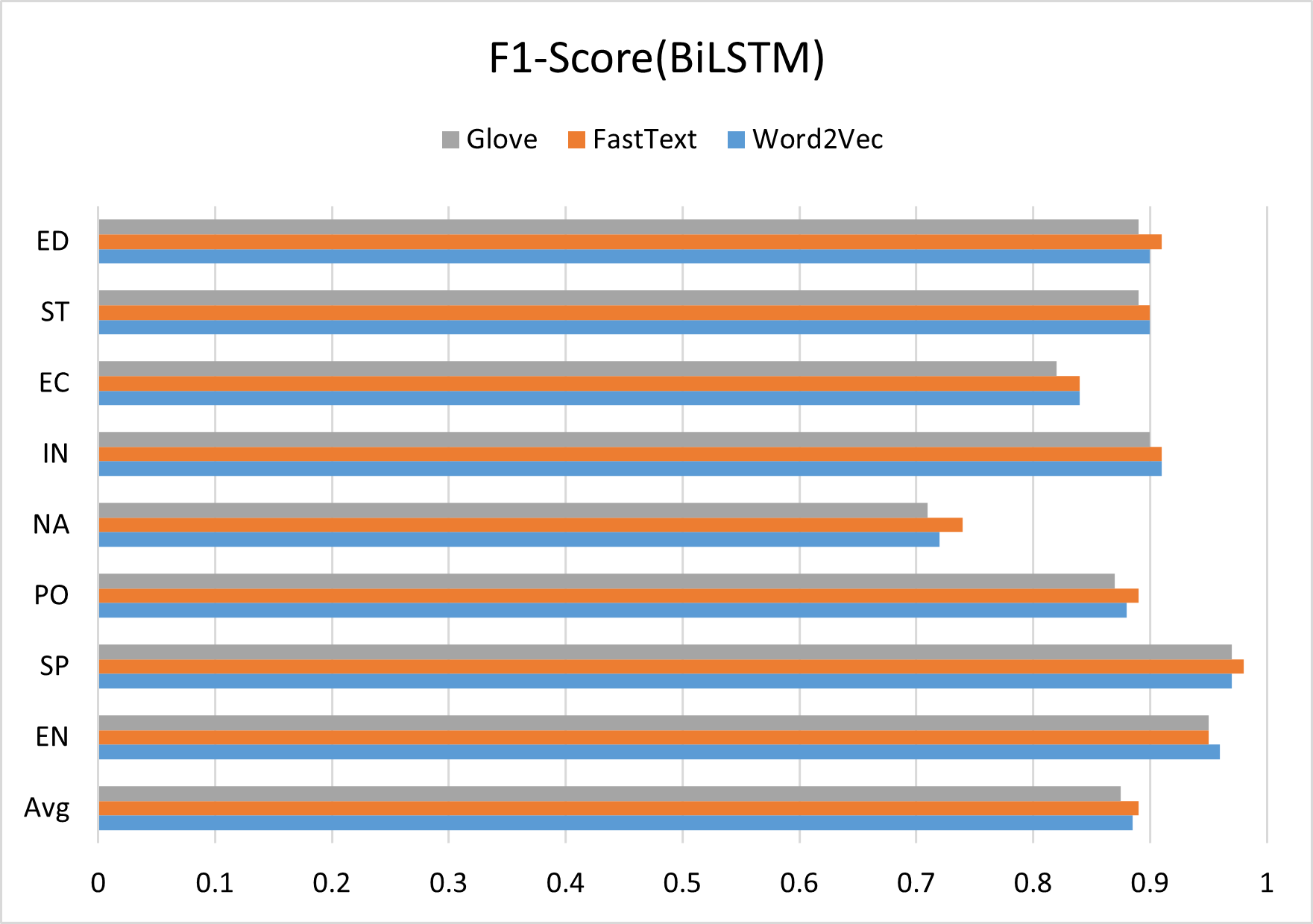}
		\captionsetup{justification=centering}
		\caption{Evaluation Metrics for Each Article Class using BiLSTM}
		\label{fig:evaluationMetrics_BiLSTM}
\end{figure}

\begin{figure}[H]
		\centering
		\includegraphics[scale=0.40]{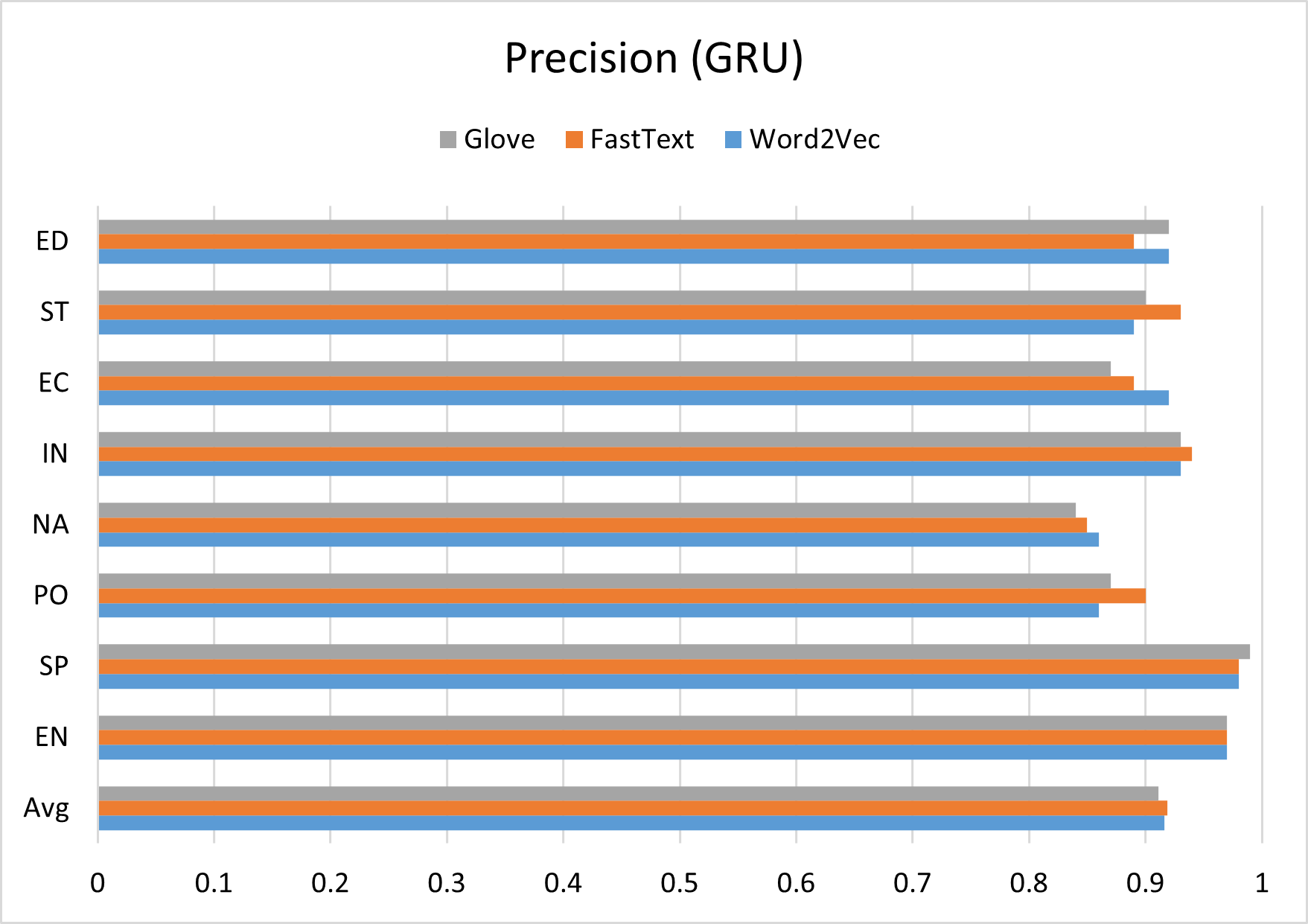}
		\includegraphics[scale=0.40]{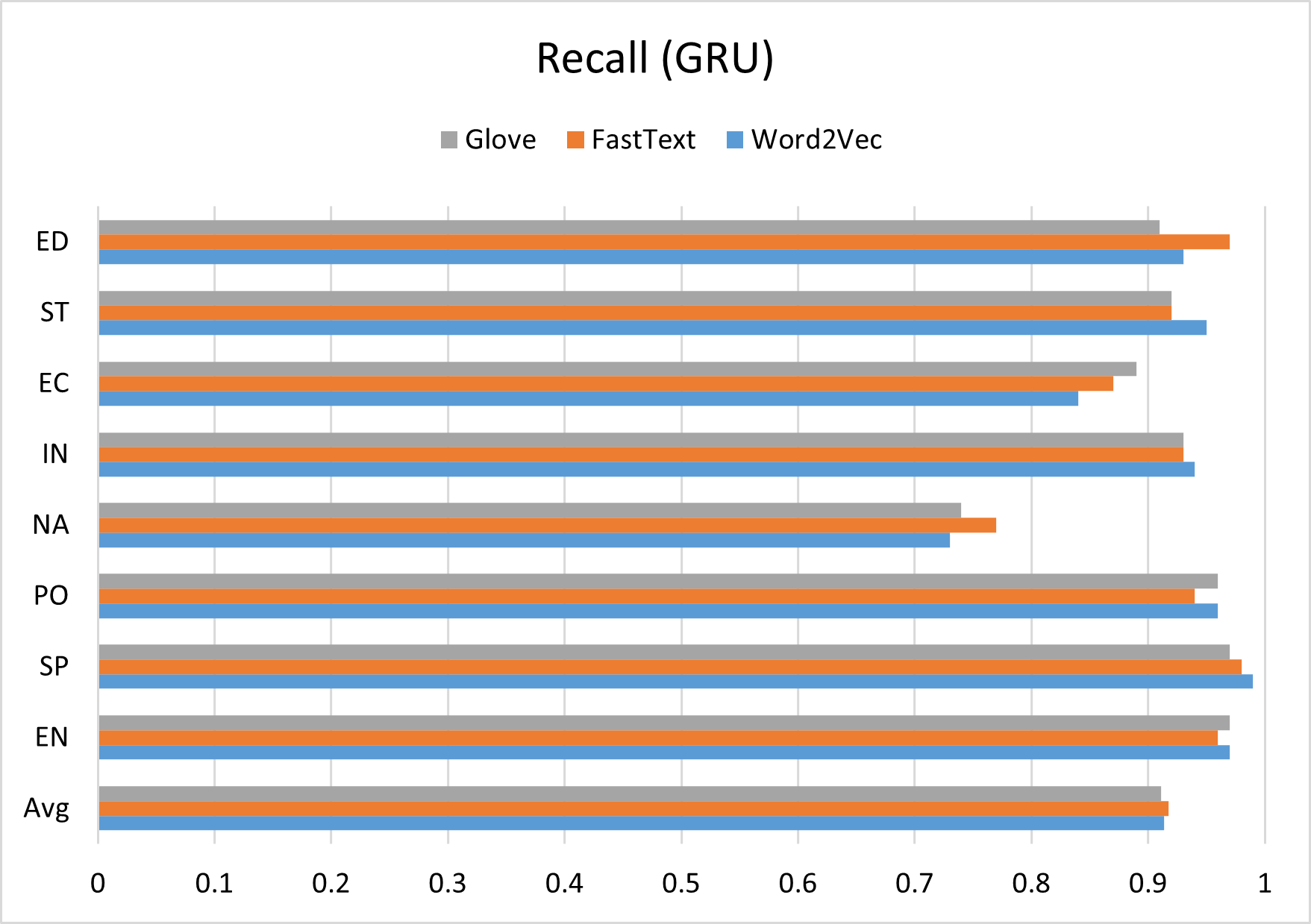}
		\includegraphics[scale=0.40]{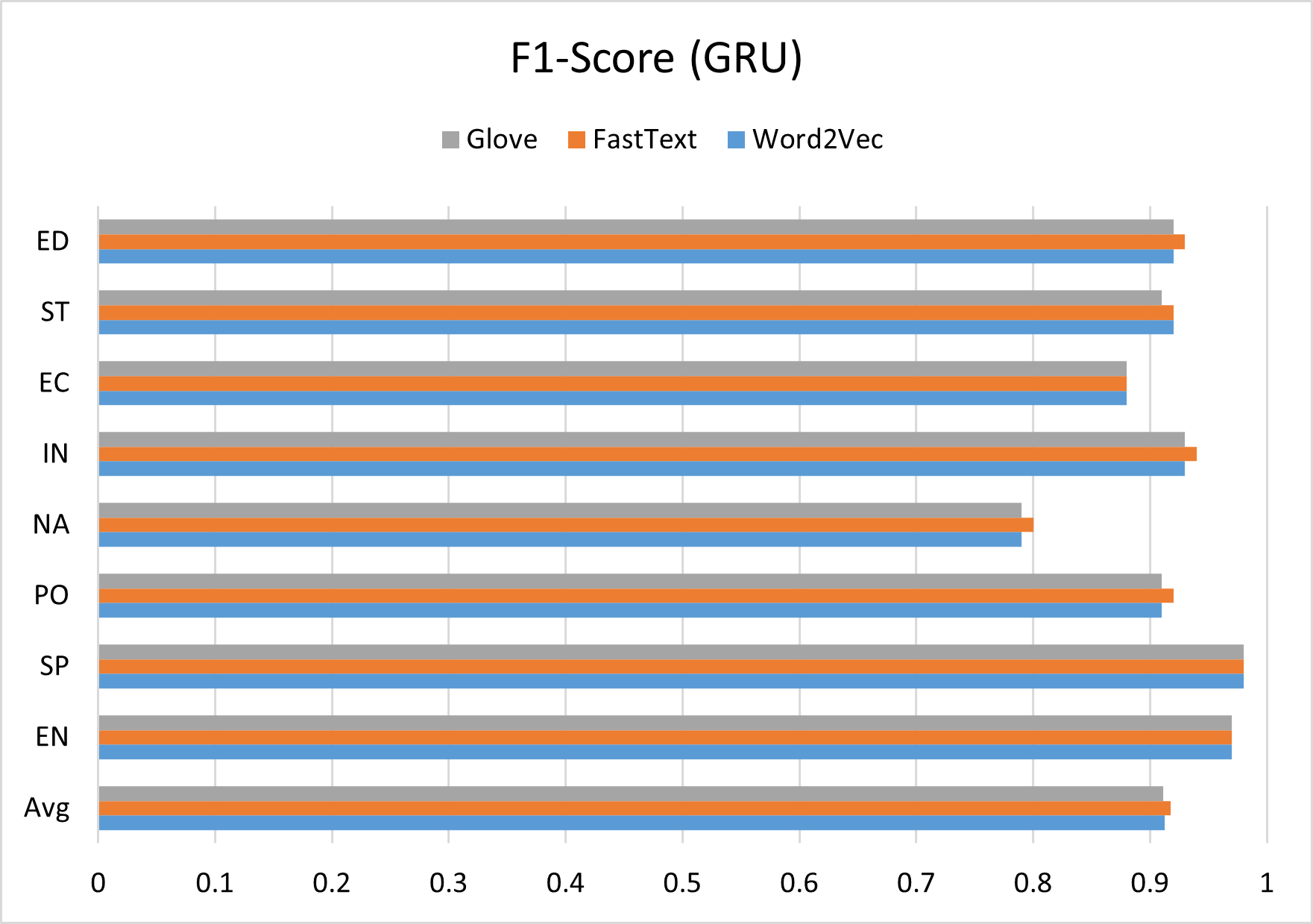}
		\captionsetup{justification=centering}
		\caption{Evaluation Metrics for Each Article Class using GRU}
		\label{fig:evaluationMetrics_GRU}
\end{figure}

Figure~\ref{fig:evaluationMetrics_CNN},~\ref{fig:evaluationMetrics_LSTM},~\ref{fig:evaluationMetrics_BiLSTM}, and~\ref{fig:evaluationMetrics_GRU} depict DL model performance for each class using f1-score, precision, and recall, as evaluation metrics. The harmonic average of recall and precision is the F1-score. The F1-score value will be 1 if both precision and recall values are 1. Precision is defined as the ratio of accurately predicted positive article classes to the total number of positively predicted classes. The recall, which is also known as the true positive rate or sensitivity, is defined as the ratio of the accurately predicted positive article classes to the total number of accurately predicted article classes. Ideally, a precision and recall value close to 1 indicates that the model has the best performance. The horizontal lines of the graph indicate the article class (see Table~\ref{tab.categoryShortform}). 

Figure~\ref{fig:evaluationMetrics_CNN} shows the CNN algorithm results (i.e., precision, recall, and F1-score) with word embedding techniques for each article class. The average precision, recall, and F1-score value for all word embedding techniques is about 0.9, demonstrating the high performance of the CNN article classification model. For the National class, the F1-score is below 0.8 for all word embedding techniques because the recall value is below 0.8, although the precision value is more than 0.8. The F1-score is about 0.9 for the remaining article classes. 

Figure~\ref{fig:evaluationMetrics_LSTM} shows the LSTM algorithm results (i.e., precision, recall, and F1-score) with word embedding techniques for each article class. The average precision, recall, and F1-score value for all word embedding techniques is approximately 0.9, presenting the high performance of the LSTM article classification model. For the National class, the F1-score is below 0.8 for all word embedding techniques because the recall and precision values are below 0.8. For the Economy class, the F1-score is below 0.9 for all word embedding techniques, as the recall and precision values are below 0.9. The F1-score, only for word2vec, is about 0.9 for the remaining article classes. The Glove and FastText do not perform well compared to Word2Vec for the LSTM article classification model. 

Figure~\ref{fig:evaluationMetrics_BiLSTM} demonstrates the BiLSTM algorithm results (i.e., precision, recall and F1-score) with word embedding techniques for each article class. The average precision, recall and F1-score value for all word embedding techniques are approximately 0.9, showing the high performance of the BiLSTM article classification model. For the National class, the F1-score is below 0.8 for all word embedding techniques because the recall and precision values are below 0.8. For the Economy, and Politics class, the F1-score is below 0.9 for all word embedding techniques. For the remaining classes, the F1-score is about 0.9, which is demonstrating high performance.

Figure~\ref{fig:evaluationMetrics_GRU} represents the GRU algorithm results (i.e., precision, recall, and F1-score) with word embedding techniques for each article class. The average precision, recall, and F1-score value for all word embedding techniques is about 0.9, presenting the heightened performance of the GRU article classification model. For the National class, the F1-score is below 0.8 for all word embedding techniques because the recall value is below 0.8, although the precision value is more than 0.8. For the Economy class, the F1-score is below 0.9 for all word embedding techniques, except the precision of Word2Vec, as the recall and precision values are below 0.9. The F1-score is about 0.9 for the remaining article classes. 

By following the above consequences, we infer that GRU with Word2Vec outperforms other DL algorithms in terms of accuracy, and evaluation metrics (i.e., precision, recall, and F1-score).

\section{Automatically labelled News Article Classification}  \label{sec.automatic.result}

This section goes through our findings related to the automatically labelled dataset in depth. We evaluate automatic single labelled dataset in two ways: first, we consider the most dominant automatic cluster label (see Section~\ref{sec.automatic.single}) as the predicted class and the original class as the tested class. After that, we evaluate the predicted and tested classes by using precision, recall, and the F1-score. Secondly, we divide the automatically labelled dataset into training and testing sets based on threshold values (i.e., 0.5, 0.6, 0.7, 0.8, and 0.9). Furthermore, we use ML algorithms to train the models and evaluate them.

\begin{figure}[H]
	\centering
	\includegraphics[scale=0.9]{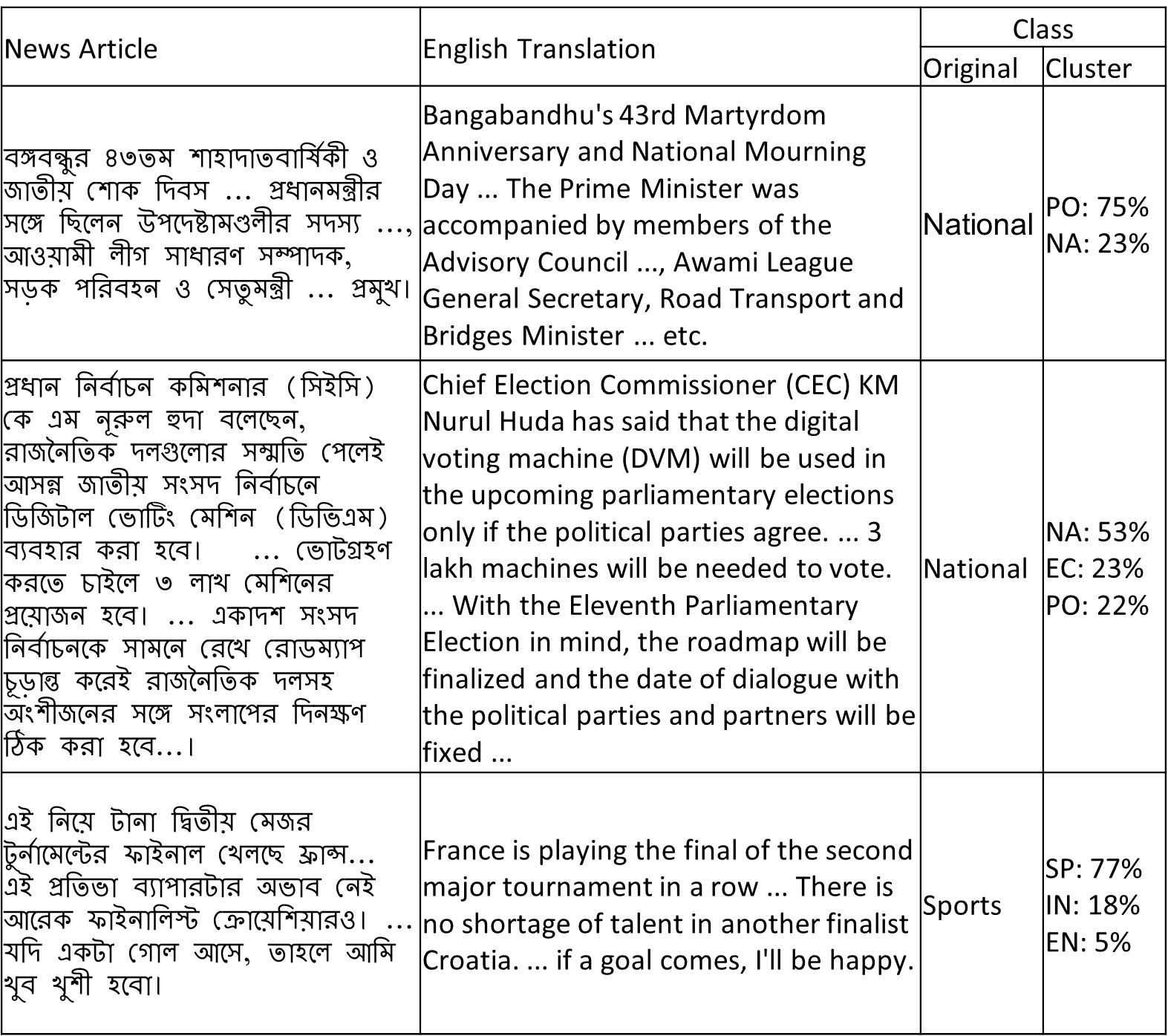}
	\captionsetup{justification=centering}
	\caption{News Article Example of Original and Cluster Class}
	\label{tab.originalvsclusterexample}
\end{figure}

Table~\ref{tab.originalvsclusterexample} lists three examples to analyse the result of automatic labelling. The first column contains the original news article text in Bengali, and the second column translates the news article into English for Bengali non-speakers to understand the text. Column 3 contains the article class, which is divided into two sub-columns: one column contains the original class, and the other column contains the cluster class. We perceive that the first example describes the opinions of several political ministers about national mourning day. This news article is mostly political and also related to national issue though it was originally labelled as National by the news reporter. After automatic labeling, we observed that this example is 75\% Politics and 23\% National. Consequently, we labelled this example as Politics based on the highest probability. The second example is describing the opinion of the election committee about the use of the digital voting machine in the election, which cost about 300k. This example is originally labelled as National and the automatic labeling marked it as 53\% National, 22\% Politics and 23\% Economic. Finally, we labelled that as National because of the highest probability. The third example is about International football news, which is originally labelled as Sports but after automatic labeling, we discover that it’s 5\% Entertainment, 18\% International, 77\% Sports and finally labelled it as Sports. After reviewing the above examples, we ascertain that single-labelled news articles can’t properly describe the news as compared to multi-labelled news articles. Consequently, we extended our research to the multi-label classification.

\begin{figure}[H]
		\centering
		\includegraphics[scale=0.70]{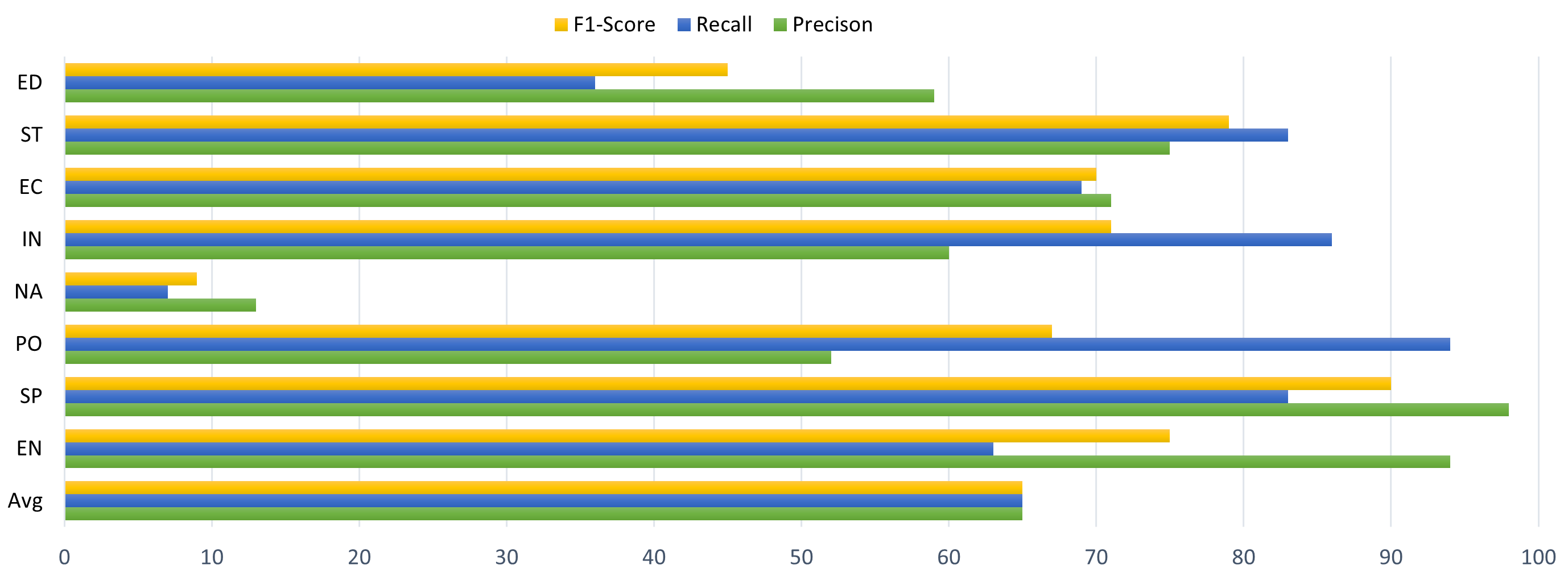}
		\captionsetup{justification=centering}
		\caption{Compare Result between Original Class and Cluster Class}
		\label{fig.originalvsclusteraccuracy}
\end{figure}

Figure~\ref{fig.originalvsclusteraccuracy} demonstrates the classification report for each class where we consider the most dominant automatic cluster label as the predicted class and the original class as the tested class. The horizontal lines of the graph indicate the article class (see Table~\ref{tab.categoryShortform}). We achieved 66\% average precision, recall, and F1-score for automatic labelling, but remarkably low precision, recall, and F1-score for the National class. As we see in the first example in Table~\ref{tab.originalvsclusterexample}, the news is more related to the Politics about 75\% and 23\% is National. The news reporter labelled this example as National although the news is more about politics. Generally, the National class can include other classes, for example, Politics, Sports, Education, and so on. As a result, our automatic labelled evaluation outcomes are low for the National class. The Education class achieved a 50\% F1-score, whereas the remaining classes achieved more than a 60\% F1-score. Apart from the National class, our automatically labelled model performs well for the remaining classes.

\begin{figure}[H]
		\centering
		\includegraphics[scale=0.9]{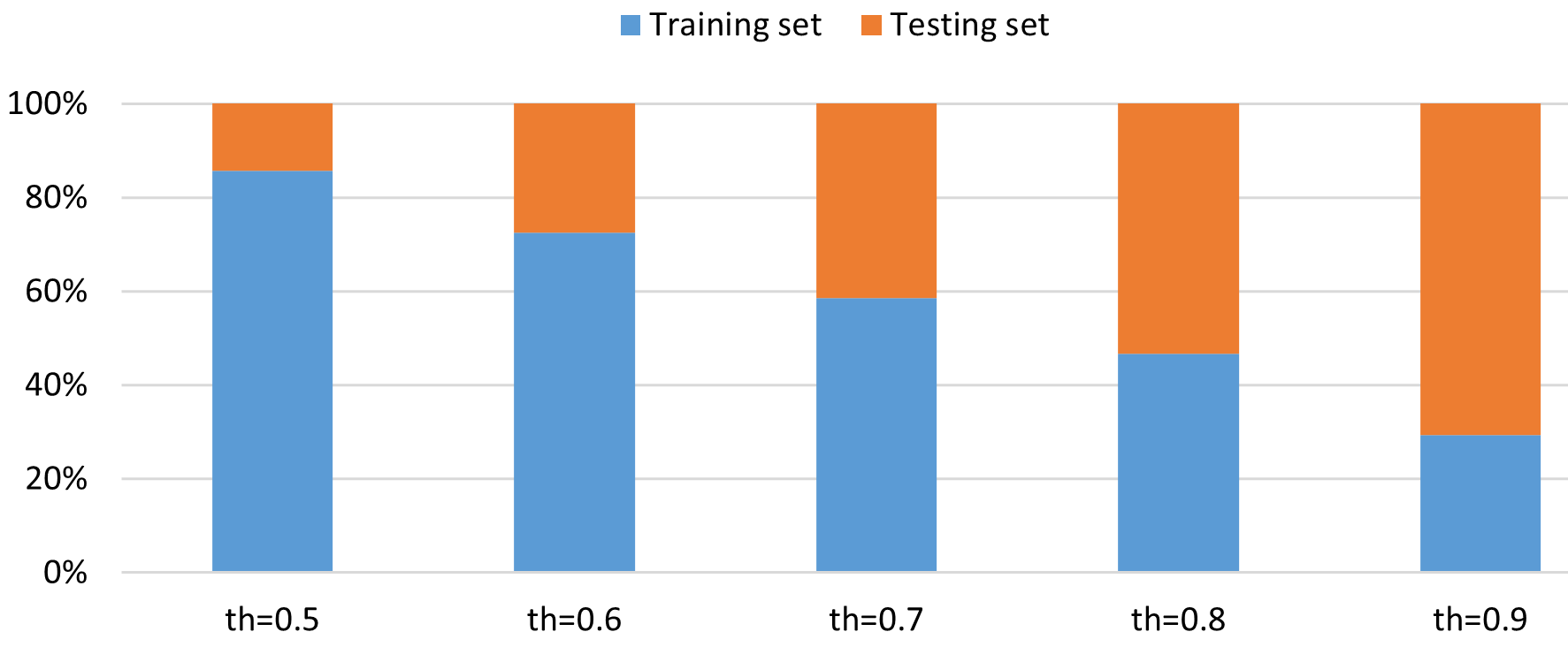}
		\captionsetup{justification=centering}
		\caption{Training and Testing Set for ML Algorithms using Topic Modeling}
		\label{fig.topic_ml}
\end{figure}

\begin{figure}[H]
		\centering
		\includegraphics[scale=0.5]{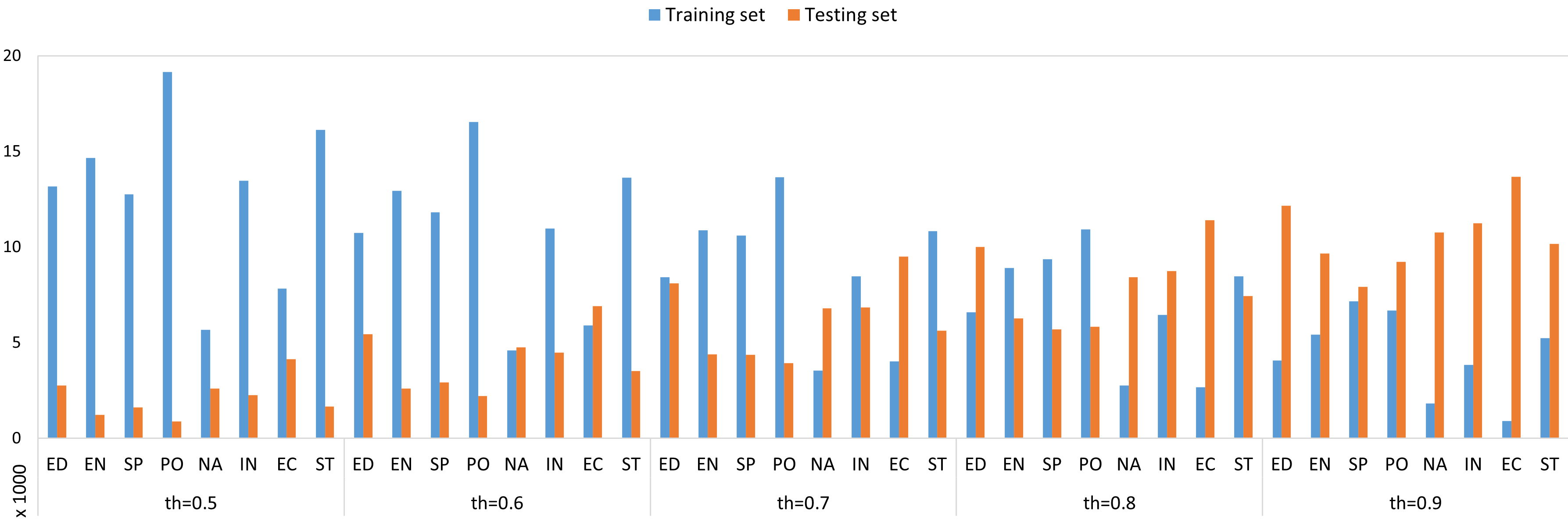}
		\captionsetup{justification=centering}
		\caption{Details of Training and Testing Set for ML Algorithms using Topic Modeling}
		\label{fig.TM_trts_details}
\end{figure}

Figure~\ref{fig.topic_ml} describes the distribution of training and testing sets for the different threshold values. More details of the training and testing set are given in Figure~\ref{fig.TM_trts_details}. For threshold 0.5, the dataset has 82\% training and 18\% testing set, furthermore, the proportion of training and testing set continuously changes for different threshold values. The training set is about 30\% and the testing set is 70\% for a threshold of 0.9. 

In Section~\ref{sec.result.manual.ml}, we achieved the highest accuracy for the Doc2Vec word embedding method using several machine learning algorithms for the originally labelled dataset. Here, we also use the Doc2Vec method for the same machine learning algorithms. We calculate the accuracy of the model using the auto-labelled test set news articles' predicted class and the manually (originally) labelled class of the following test set news articles.

\begin{figure}[H]
		\centering
		\includegraphics[scale=0.70]{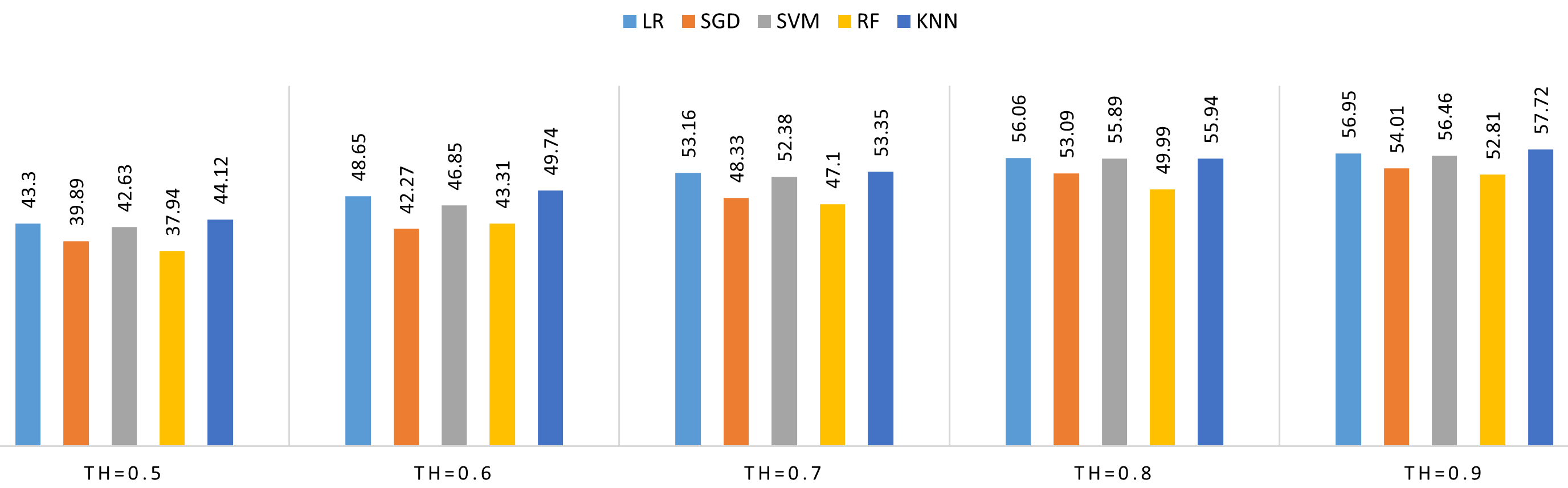}
		\captionsetup{justification=centering}
		\caption{Result of ML Algorithms using Topic Modeling}
		\label{fig.TM_result_ML}
\end{figure}

Figure~\ref{fig.TM_result_ML} shows the accuracy of several machine learning algorithms with different threshold values. The highest accuracy of 57.72\% was achieved by the KNN algorithm for threshold 0.9.

\begin{figure}[H]
		\centering
		\includegraphics[scale=0.6]{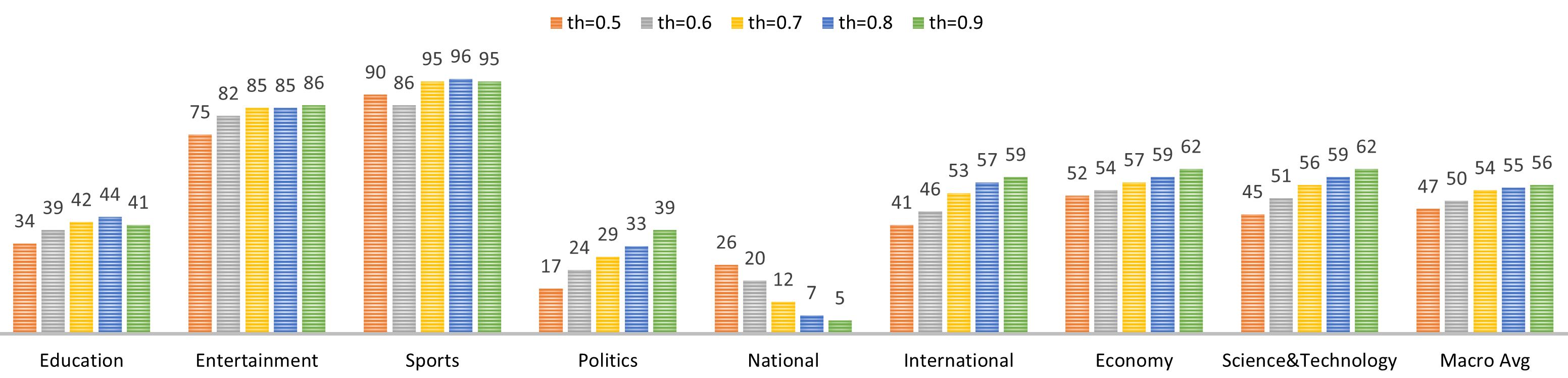}
		\captionsetup{justification=centering}
		\caption{Precision of ML Algorithms using Topic Modeling}
		\label{fig.TM_precision}
\end{figure}

\begin{figure}[H]
		\centering
		\includegraphics[scale=0.6]{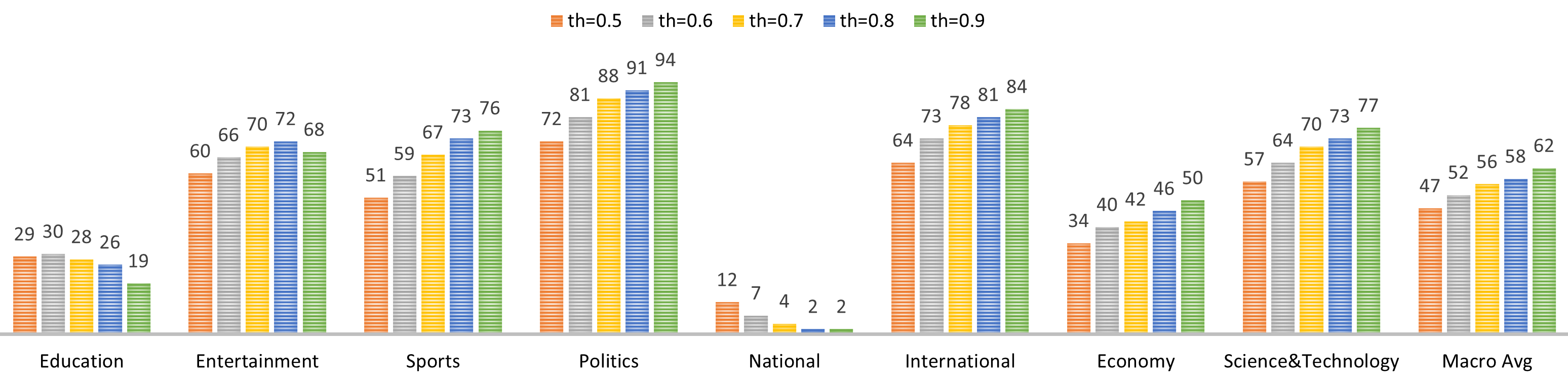}
		\captionsetup{justification=centering}
		\caption{Recall of ML Algorithms using Topic Modeling}
		\label{fig.TM_recall}
\end{figure}

\begin{figure}[H]
		\centering
		\includegraphics[scale=0.6]{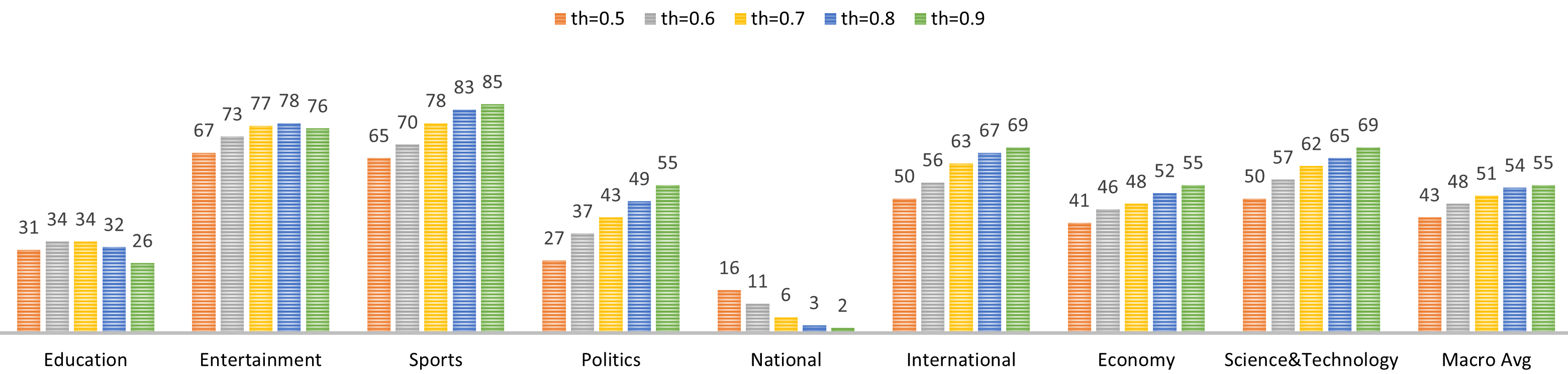}
		\captionsetup{justification=centering}
		\caption{F1-Score of ML Algorithms using Topic Modeling}
		\label{fig.TM_f1score}
\end{figure}

The precision, recall, and f1-score for each class using KNN algorithms are shown in Figures~\ref{fig.TM_precision},~\ref{fig.TM_recall}, and~\ref{fig.TM_f1score} at various threshold values. The automatically labelled dataset performed better for almost all classes excluding National and Education. As we see in Table~\ref{tab.originalvsclusterexample}, some news articles are labelled as National, although they are strongly or partially related to other topics.



We evaluated automated single-labelled news articles and concluded that a single label is insufficient to accurately classify a news article. As a consequence, we look into the performance of multi-label news article classification in this section. The evaluation of multi-label news classification is more difficult than that of single-label text classification. The multi-label evaluation metrics are discussed further below. 

Let a multi-label dataset T has a label set $L, |L| = k$, and n multi-label instances $(x_i, Y_i), 1 \leq i \leq n, (x_i \epsilon X , Y_i \epsilon y = \left\{0, 1\right\}^k)$. The multi-label classifier h and $Z_i =h(x_i) = \left\{0, 1 \right\}^k$ be the set of predicted labels by h for the instance $x_i$~\cite{sorower2010literature}.

Accuracy (A): The percentage of predicted accurate labels to the total number (predicted and actual) of labels for each occurrence is defined as accuracy. The overall accuracy is calculated as the average of all cases.
\begin{equation}
Accuracy, A = \frac{1}{n} \sum_{i=1}^{n} \frac{|Y_{i} \bigcap Z_{i}|} {|Y_{i} \bigcup Z_{i}|}
\end{equation}
Precision (P): Precision is defined as the ratio of expected accurate labels to total number of actual labels, averaged over all cases.
\begin{equation}
Precision, P = \frac{1}{n} \sum_{i=1}^{n} \frac{|Y_{i} \bigcap Z_{i}|} {|Z_{i}|}
\end{equation}

Recall (R): The dimension of correctly predicted labels to the cumulative number of predicted labels, averaged over all examples, is referred to as recall. 

\begin{equation}
Recall, R = \frac{1}{n} \sum_{i=1}^{n} \frac{|Y_{i} \bigcap Z_{i}|} {|Y_{i}|} 
\end{equation}

F1-Measure (F): It is the harmonic mean of precision and recall.

\begin{equation}
F1 = \frac{1}{n} \sum_{i=1}^{n} \frac{2|Y_{i} \bigcap Z_{i}|} {|Y_{i}| + |Z_{i}|}
\end{equation}

Hamming Loss (HL): The Hamming Loss statistic calculates the average number of times an example's relevance to a class label is predicted incorrectly. As a consequence, hamming loss takes into account both the prediction error (when an incorrect label is predicted) and the missing error (when a relevant label is not predicted), standardized across all classes and instances.

\begin{equation}
Hamming Loss, HL = \frac{1}{kn} \sum_{i=1}^{n} \sum_{l=1}^{k} [I(l \epsilon Z_{i} \wedge l \notin Y_{i}) + I(l \notin Z_{i} \wedge l \epsilon Y_{i})]
\end{equation}

where the indicator function is ‘I’. In terms of multi-label classification model performance, the less the hamming loss, the better the learning algorithm performs. The range of humming loss is between 0 to 1.

\begin{figure}[H]
		\centering
		\includegraphics[scale=0.7]{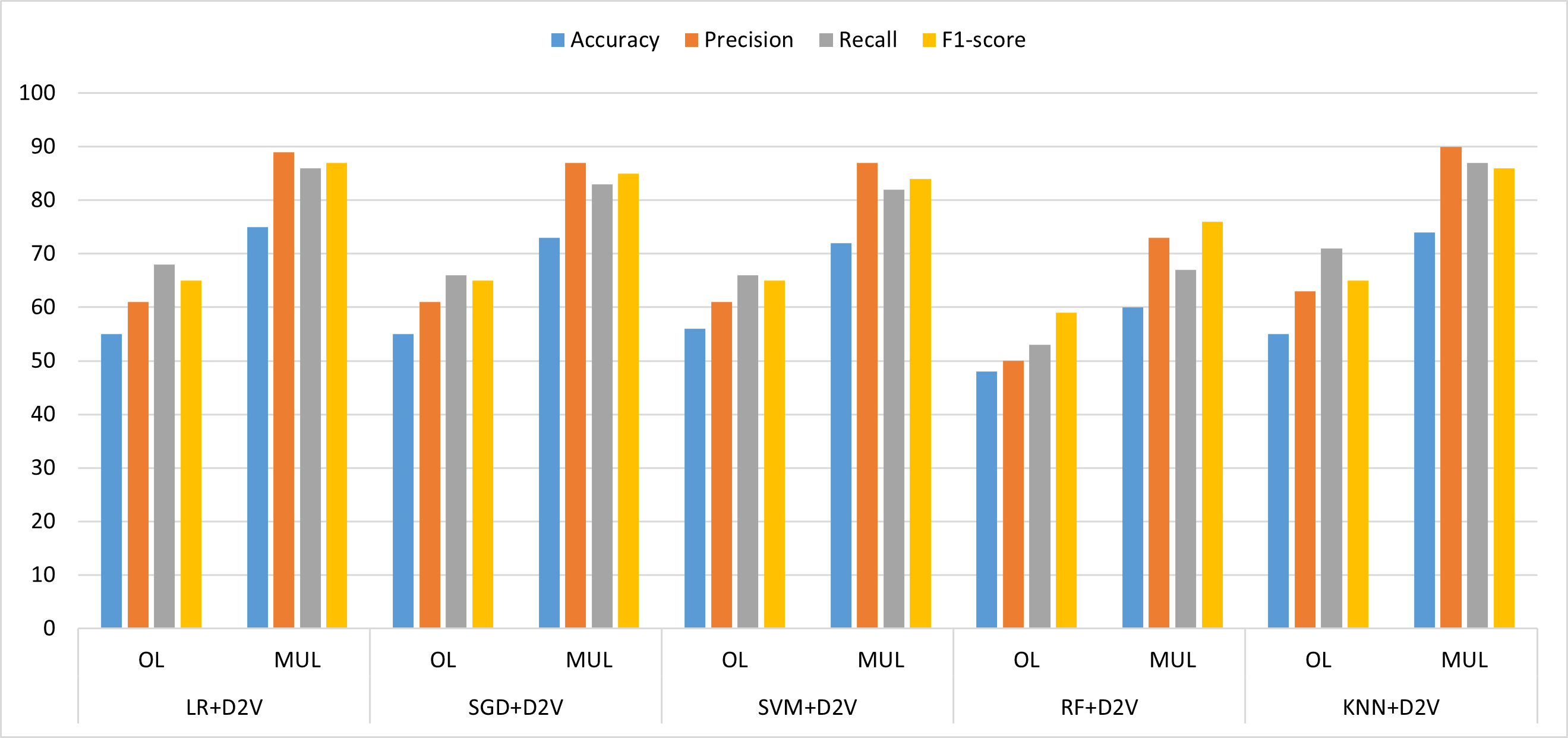}
		\captionsetup{justification=centering}
		\caption{Comparison of Single Label and Multi-label Text Classification}
		\label{fig.multilabel_result}
\end{figure}

The multi-label classification was performed using the same ML algorithms and the Doc2Vec word embedding method (see Section~\ref{sec.automatic.multi}), which has been shown to be effective for single-label classification. We use a minimum threshold value of 0.3 for multi-labelling. We evaluate the multi-label news classification model in two testing set: using the original label, and predicted multi-label. The original label is a single label that cannot be evaluated with a multi-label evaluation process. As a result, we transform the single label into a multi-label by assigning 1 to the original label and 0 to the remainder of the label. For example, in multi-label, we have 8 labels for 8 classes and in the single label, we have only one label. Consequently, we assign 1 to the original label, and 0 to the remaining label, like this (1,0,0,0,0,0,0,0). Figure~\ref{fig.multilabel_result} shows the multi-label classification evaluation metrics. The OL refers to the original label and MUL refers to automatic multi-label. The highest accuracy, about 75\%, is achieved by the KNN algorithm. It also performs the best with 90\% precision, 88\% recall, and 87\% F1-score.

\begin{figure}[H]
		\centering
		\includegraphics[scale=0.6]{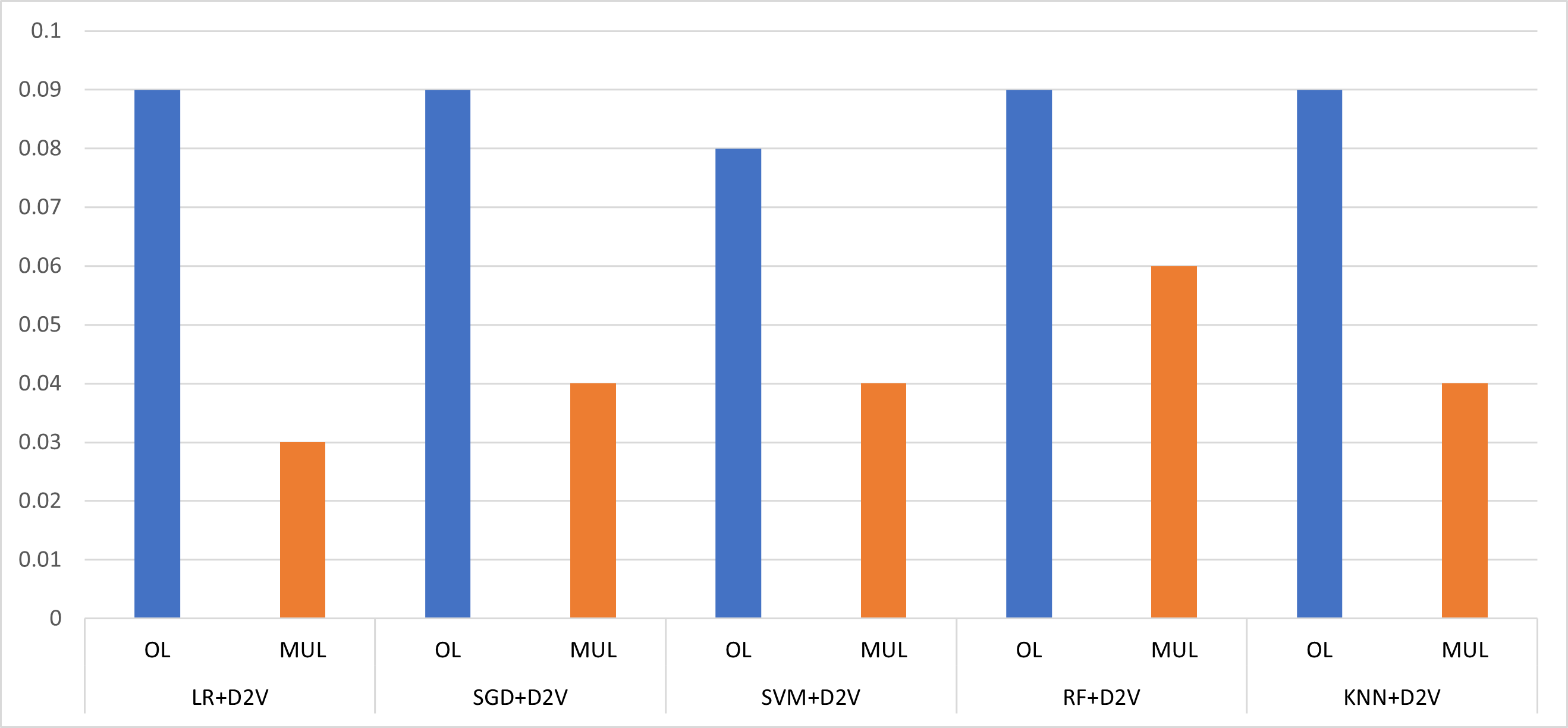}
		\captionsetup{justification=centering}
		\caption{Comparison of Single Label and Multi-label Text Classification using Hamming Loss}
		\label{fig.multilabel_resultHL}
\end{figure}

Figure \ref{fig.multilabel_resultHL} shows that both testing sets (original label class and multi-label class) have a very low hamming loss of 0.09 and 0.03 for logistics regression, respectively. 

\section{Discussion}\label{sec.discussion}

The broader aim of our work is to investigate how to name the parameter with the best output. In this paper, we investigate the following studies: we develop a comprehensive Bangla news article dataset using web scraping technique~\cite{https://doi.org/10.48550/arxiv.2210.09389}; present a case study to investigate the performance of machine learning and deep learning algorithms with various word embedding techniques(see Section~\ref{sec.methodology}, and~\ref{sec.result.manual.ml}); propose a method for developing automatically labelled datasets, i.e., single-label and multi-label, from manually labelled datasets(see Section~\ref{sec.methodology}), and investigate automatic labelling(see Section~\ref{sec.automatic.result}). Automatic labelling is one of the issues which is addressed in Deep Journalism and DeepJournal V1.0~\cite{ahmad2022deep}.  

For news article classification, we used a manually labelled Potrika dataset, which is labelled by the news reporter. We analysed the performance of word embedding techniques (i.e., BOW, TF-IDF, Doc2vec, word2Vec, fasttext, and glove), machine learning (i.e., logistic regression, SGD, SVM, random forest, and KNN), and deep learning algorithms (CNN, LSTM, BiLSTM, and GRU). We conclude that the deep learning algorithm's performance is much better compared to machine learning algorithms as deep learning algorithms hold the contextual meaning of the articles. On the other hand, machine learning algorithms are based on statistical learning theory and are unable to capture the context of the articles. Word embedding techniques, such as BOW and TF-IDF, achieved very low accuracy, whereas Doc2vec got good accuracy of 87.14\% for logistic regression because the Doc2vec model is performing as a memory that recalls what is lacking from the present context. Deep learning algorithms achieved high accuracy for all word embedding techniques. Additionally, we investigate the performance of deep learning algorithms in two ways: using a stemmer and without a stemmer. We discovered that the use of stemmer has no significant impact in terms of the performance of the algorithms. GRU achieved the highest accuracy of about 92\% using the fasttext word embedding technique. 


\section{Conclusion and Future Work} \label{sec.conclusion}
In this research, we have used the most comprehensive Bangla newspaper dataset called Potrika and performed extended experimentation to compare the performance of several machine learning and deep learning algorithms using several word embedding techniques. We further investigate the possibility of using topic modeling algorithms for automatic labeling of the classification dataset. Besides, we evaluate the multi-label news article classification with the state-of-the-art evaluation metrics. In the future, we will adopt hybrid deep learning algorithms with several word embedding techniques to improve the accuracy of Bangla news article classification and also improve the performance of automatic labeling techniques.


\section*{Acknowledgments}
The work reported in this paper is supported by the High Performance Computing Centre (HPC Center) at King Abdulaziz University, Saudi Arabia. The experiments reported in this paper were performed on the Aziz supercomputer at the HPC Center, King Abdulaziz University.

\section*{Declaration of Competing Interest}
The authors declare that they have no known competing financial interests or personal relationships that could have appeared to influence the work reported in this paper.

\bibliographystyle{unsrt}  
\bibliography{20210128_Text_Classification_IstiakAhmad_V24_RMv3}

\end{document}